\documentclass[journal,onecolumn]{IEEEtran}
\usepackage{graphicx}
\usepackage{stfloats}
\usepackage{mathrsfs}  
\usepackage{lipsum}
\usepackage{amsmath}
\usepackage{bm}
\usepackage{algorithm}
\usepackage{algpseudocode}

\ifCLASSINFOpdf
\else
\fi
\begin{document}
%
% paper title
% Titles are generally capitalized except for words such as a, an, and, as,
% at, but, by, for, in, nor, of, on, or, the, to and up, which are usually
% not capitalized unless they are the first or last word of the title.
% Linebreaks \\ can be used within to get better formatting as desired.
% Do not put math or special symbols in the title.
\title{SUGAMAN: Describing Floor Plans for Visually Impaired by Annotation Learning and Proximity based Grammar}
%
%
% author names and IEEE memberships
% note positions of commas and nonbreaking spaces ( ~ ) LaTeX will not break
% a structure at a ~ so this keeps an author's name from being broken across
% two lines.
% use \thanks{} to gain access to the first footnote area
% a separate \thanks must be used for each paragraph as LaTeX2e's \thanks
% was not built to handle multiple paragraphs
%

\author{Shreya Goyal, Satya Bhavsar,  Shreya Patel, Chiranjoy Chattopadhyay, Gaurav Bhatnagar}% <-this % stops a space

% note the % following the last \IEEEmembership and also \thanks - 
% these prevent an unwanted space from occurring between the last author name
% and the end of the author line. i.e., if you had this:
% 
% \author{....lastname \thanks{...} \thanks{...} }
%                     ^------------^------------^----Do not want these spaces!
%
% a space would be appended to the last name and could cause every name on that
% line to be shifted left slightly. This is one of those "LaTeX things". For
% instance, "\textbf{A} \textbf{B}" will typeset as "A B" not "AB". To get
% "AB" then you have to do: "\textbf{A}\textbf{B}"
% \thanks is no different in this regard, so shield the last } of each \thanks
% that ends a line with a % and do not let a space in before the next \thanks.
% Spaces after \IEEEmembership other than the last one are OK (and needed) as
% you are supposed to have spaces between the names. For what it is worth,
% this is a minor point as most people would not even notice if the said evil
% space somehow managed to creep in.

% The paper headers
\markboth{Journal of \LaTeX\ Class Files,~Vol.~14, No.~8, August~2015}%
{Shell \MakeLowercase{\textit{et al.}}: Bare Demo of IEEEtran.cls for IEEE Journals}
% The only time the second header will appear is for the odd numbered pages
% after the title page when using the twoside option.
% 
% *** Note that you probably will NOT want to include the author's ***
% *** name in the headers of peer review papers.                   ***
% You can use \ifCLASSOPTIONpeerreview for conditional compilation here if
% you desire.

% If you want to put a publisher's ID mark on the page you can do it like
% this:
%\IEEEpubid{0000--0000/00\$00.00~\copyright~2015 IEEE}
% Remember, if you use this you must call \IEEEpubidadjcol in the second
% column for its text to clear the IEEEpubid mark.

% use for special paper notices
%\IEEEspecialpapernotice{(Invited Paper)}

% make the title area
\maketitle

% As a general rule, do not put math, special symbols or citations
% in the abstract or keywords.
\begin{abstract}
In this paper, we propose SUGAMAN (Supervised and Unified framework using Grammar and Annotation Model for Access and Navigation). SUGAMAN is a Hindi word meaning ``easy passage from one place to another''. SUGAMAN synthesizes textual description from a given floor plan image for the visually impaired. A visually impaired person can navigate in an indoor environment using the textual description generated by SUGAMAN. With the help of a text reader software the target user can understand the rooms within the building and arrangement of furniture to navigate. SUGAMAN is the first framework for describing a floor plan and giving direction for obstacle-free movement within a building. We learn $5$ classes of room categories from $1355$ room image samples under a supervised learning paradigm. These learned annotations are fed into a description synthesis framework to yield a holistic description of a floor plan image. We demonstrate the performance of various supervised classifiers on room learning. We also provide a comparative analysis of system generated and human written descriptions. SUGAMAN gives state of the art performance on challenging, real-world floor plan images. This work can be applied to areas like understanding floor plans of historical monuments, stability analysis of buildings, and retrieval.
\end{abstract}

% Note that keywords are not normally used for peerreview papers.
%\begin{IEEEkeywords}
%IEEE, IEEEtran, journal, \LaTeX, paper, template.
%\end{IEEEkeywords}

% For peer review papers, you can put extra information on the cover
% page as needed:
% \ifCLASSOPTIONpeerreview
% \begin{center} \bfseries EDICS Category: 3-BBND \end{center}
% \fi
%
% For peerreview papers, this IEEEtran command inserts a page break and
% creates the second title. It will be ignored for other modes.
\IEEEpeerreviewmaketitle

\section{Introduction}
% The very first letter is a 2 line initial drop letter followed
% by the rest of the first word in caps.
% 
% form to use if the first word consists of a single letter:
% \IEEEPARstart{A}{demo} file is ....
% 
% form to use if you need the single drop letter followed by
% normal text (unknown if ever used by the IEEE):
% \IEEEPARstart{A}{}demo file is ....
% 
% Some journals put the first two words in caps:
% \IEEEPARstart{T}{his demo} file is ....
% 
% Here we have the typical use of a "T" for an initial drop letter
% and "HIS" in caps to complete the first word.
One of the primary goals of a combined framework involving computer vision (CV) and natural language processing (NLP) is to understand an image and describe it. The techniques available in CV and digital image processing (DIP) helps to localize the objects in the image, identify key attributes and provide a relationship among them. On the other hand, NLP provides us with an end to end description of that image, and thereby connecting the output of CV with the text. For the purpose of having a better understanding of a floor plan image, we propose SUGAMAN (Supervised and Unified framework using Grammar and Annotation Model for Access and Navigation), which is an attempt to connect the these two modalities, i.e. CV and NLP, in the context of document images and text data. SUGAMAN is a Hindi word which translates to easy passage from one place to another. Apart from describing the general information about the floor plan images, it also generates room to room navigation information, while avoiding the obstacles. This navigation information can be very useful for visually impaired people as it becomes difficult for them to move in an indoor environment. It will be really helpful for them if there is a system that tells about the surroundings environment in natural language. SUGAMAN generates such natural language description of an indoor environment from building floor plan images, which gives a detailed idea of the indoor environment. Here the input is a building floor plan image and the output is a textual description of the same. The description includes detail about the (i) rooms, (ii) connectivity among the rooms, (iii) type of decor within every room, and (iv) their relative position, and (v) navigational information, while avoiding obstacles. 

In the past few years, unconventional documents like floor plans, engineering drawing gained a lot of attention from the document analysis and recognition (DAR) community. Engineering drawings contains many symbols, texts, line drawings, which need to be recognized and understood. Understanding of circuit diagrams, floor plan images, machine designs, building diagrams, maps etc. has its own importance. While researchers have looked into the problems of segmentation, symbol spotting, text recognition, symbol classification etc., the problem of narration synthesis from those documents was overlooked. Generating a textual description (narration) from an engineering drawing document image can be very useful for a layman to understand the document. As a particular example, building floor plan images involve a lot of technicalities such as symbols, dimension of the building, its design, orientation, as well as the interior decors, which are not known by common users. Various techniques have been proposed for its wall segmentation, text, room and decor segmentation in order to understand them. Even though text synthesis has been an area covered under natural language processing for years and also explored with real world images, document images were always ignored.

\begin{figure}
		\centering
		\includegraphics[scale=.35]{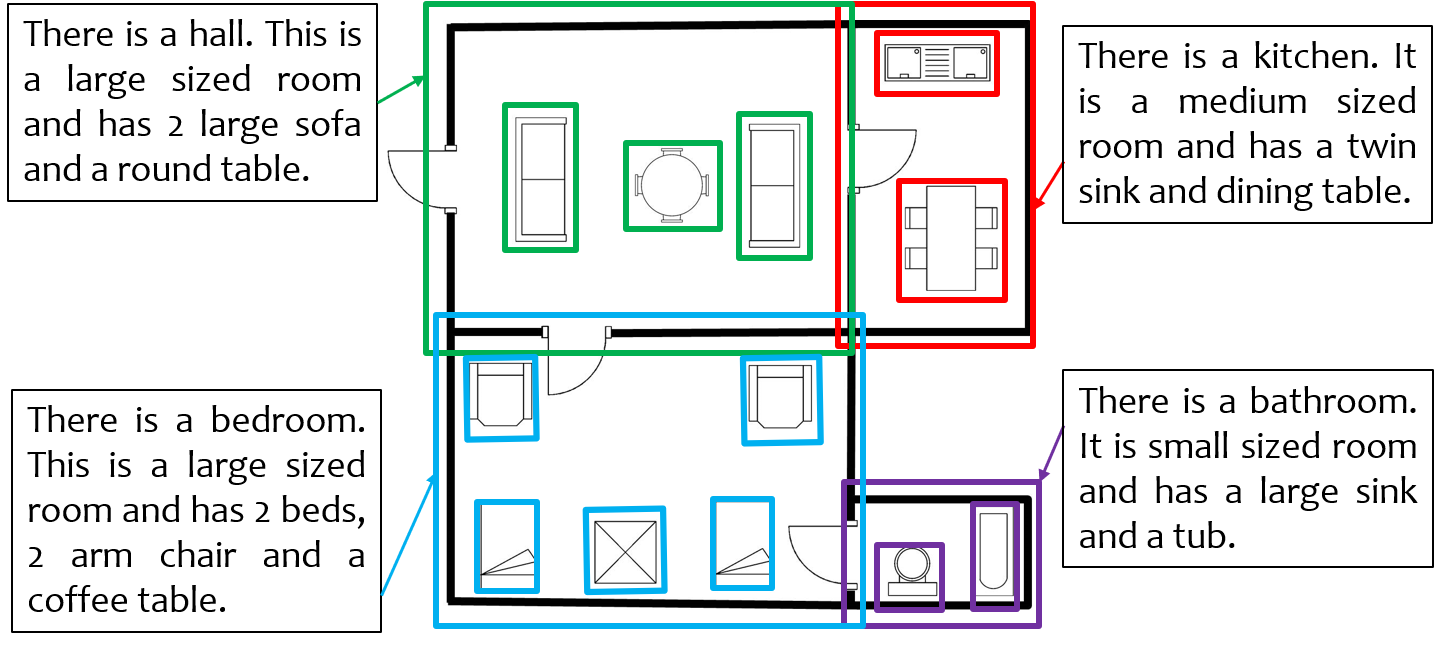}
		\caption{Introducing the problem of narration synthesis from a given input floor plan image.}
		\label{fig:problem}
	\end{figure}
	
	 Figure \ref{fig:problem} exemplifies the problem and the potential solution for a real-world floor plan images. The key characteristics that makes this work unique are: (i) proposing a unified framework for narration synthesis from floor plan images, (ii) improvement in the previously available techniques for decor characterization, (iii) proposal of a novel feature to represent a room within a floor plan, (iv) learning the room annotations for room classification, and (v) augmentation of an publicly available dataset by annotating floor plan images with textual descriptions.
	 
	 The paper is organized in the following manner: in Sec. \ref{sec:litserv} we discuss the known results that are published in the literature and are related to our framework. Overview of the proposed system is given in Sec. \ref{briefdescription}. Description about the dataset used is given in Sec. \ref{sec:dataset}. Room annotation learning and classification framework is presented in Sec. \ref{sec:RC}. The method of description synthesis is discussed in Sec. \ref{sec:description}. Results of the intermediate processing stage is discussed in Sec. \ref{sec:exp}. Results of description synthesis is given in Sec. \ref{sec:res}. Finally, the paper is concluded in Sec. \ref{sec:conclusion}.

%\subsection{Subsection Heading Here}
%Subsection text here.

% needed in second column of first page if using \IEEEpubid
%\IEEEpubidadjcol

\section{Related Work}
\label{sec:litserv}

In this section, first we describe works related to the document image processing and recognition tasks, followed by the ones proposed for description synthesis.

\subsection{Document image processing and recognition}
Symbol spotting, character recognition and feature extraction has been an evergreen research topic for classification of document images. An overview of graphics symbol recognition can be found in \cite{chhabra1997graphic}. In this matter \cite{rusinol2010state} provides a detailed state of the art survey for symbol spotting techniques and there performance analysis. 
 In \cite{yu1994isolating}, a class of drawings are analyzed which includes flow charts by pre-processing scanned binary images, symbols and connection lines followed by extraction of features like shapes, orientation of lines and connections. Trier et al. \cite{joseph1992knowledge} have also imparted knowledge based systems for interpreting symbols in engineering drawings. In \cite{lai1991detection}, with respect to engineering drawings, an approach for detecting and classifying dashed line segments was proposed. In \cite{lu1998detection}, a new rule based algorithm was proposed to differentiate between text and graphics part from engineering drawings.  
  
In a floor plan, the decors are an inseparable component and visually similar to characters. Properties of characters and their recognition methods could be adopted for recognizing decors. The authors of \cite{trier1995data} used topological analysis method to recognize characters and lines from gray scale scanned images of hand printed documents, whose information has been lost after binarization. In \cite{fletcher1988robust}, CCA and histogram based thresholding is applied to separate text and graphics and Hough transform is used to group together components into logical strings, which was later improved by Tombre et al. \cite{tombre2002text} through right choice of threshold and its stability. 

Language invariant script recognition was proposed in \cite{spitz1997determination} using spatial relationships of features. Two classes of languages are determined using optical density distribution and most frequently occurring word shapes characteristics.
Decors symbols can also be considered as line drawing images. In that context Freeman \cite{freeman1974computer} discusses various forms of line drawing representations, processing of line drawing structures derived from images by extracting their features like chain coding scheme, polygon approximation. Qureshi et al. \cite{qureshi2007spotting} have proposed a solution for symbol spotting using a graph representation of graphical documents. In the same line Dutta et al. \cite{dutta2013symbol} has proposed a symbol spotting technique in graphical documents, in which graph represents the document and a sub-graph matching is used to spot symbols. Viola and Jones \cite{viola2001rapid} proposed a learning algorithm by selecting salient visual features and a novel representation called ``integral image'',  which allows the features used by detector to be computed very quickly. They combined increasingly more complex classifiers in a ``cascade'', which allows background regions of the image to be quickly discarded speeding up object detection. \cite{lienhart2002extended}, proposed a new rotated Haar like features was proposed, which improves the accuracy of object detection by great extent. In \cite{joachims1998text},Joachims discussed about bag of words approach for objects categorization. 
 However in another work \cite{yadav2016text}, an image is divided into blocks, and blocks with higher density are considered as text by grouping them, where key points are extracted by using FAST \cite{rosten2010faster} method.

 In the context of floor plans text/graphic segmentation were performed in \cite{ahmed2011text,ahmed2011improved,ahmed2012automatic}. In \cite{mello2012automatic}, a new algorithm to segment the ancient maps and floor plans was proposed by removing non textual elements and recognizing characters to identify the plans. In  \cite{ahmed2012automatic}, various rooms are detected and labelled by Optical Character Recognition (OCR) on localized text regions. Heras et al. \cite{de2014statistical} have proposed a Statistical patch based Bag of visual words (BoVW) model to segment floor plan image. Attributed graph of line segments is generated with nodes labelled for wall segmentation, followed by DFS search to obtain walls in \cite{de2014statistical}. In \cite{de2014statistical}, doors and windows are detected using symbol spotting techniques using SURF feature by detecting the key points in the image. In \cite{mace2010system}, walls in a floor plan image are detected by exploiting the properties of Hough transform on vectorized image. Also they have used bag of visual words for the same task and later A* search is performed to find a path between doors. In \cite{de2014statistical}, rooms are detected in the floor plan images by finding the closed regions in WDWC graph, preceded by removal of all the terminals of the graph. Wall contained image is decomposed into convex regions to detect the rooms and holes in the polygons are resolved. Later convex regions are checked for over segmentation by identifying the fitting rectangle of that region in \cite{mace2010system}. In \cite{ojala2002multiresolution}, a method for recognizing ``uniform'' binary patterns was proposed.

	\subsection{Image description generation}
	Description generation from images has been an interesting area for aligning images to their corresponding text. Although literature is available on natural images, description of document images is still an untouched domain. In one of our earlier work \cite{shreya2018} we have introduced the problem of description generation from floor plan. An extensive survey of the existing techniques is given in \cite{bernardi2016automatic}. Vinyals et al. \cite{vinyals2015show} presented a generative model, based on deep recurrent architecture, which takes an input image $I$, and trained to maximize the likelihood $p(S|I)$ of producing a target sequence of words $S = {S_1, S_2,\dots}$. In the same line \cite{xu2015show} discusses a model to generate descriptions from images, which automatically learns to describe the content of images. The work in \cite{kuznetsova2012collective} have presented a holistic data-driven approach to image description generation, which uses vast amount of image data and associated description available over internet. They recognize and predict the contents of an image and then use existing human composed description to generate natural captions for images.  In \cite{karpathy2015deep,karpathy2014deep}, authors have generated dense description of images, by developing a deep neural network model. They introduced a multimodal Recurrent Neural Network architecture that takes an input image and generates its description in text. In \cite{karlsen2013automatic}, authors have used image meta-data for automatically generating textual image collection descriptions that include both image content and context information. Moreover, they convert and expand the meta-data, by using publicly available information and services over internet. Kulkarni et al. \cite{kulkarni2011baby} generated description of images by detecting objects, modifiers (adjectives) and spatial relationships (prepositions) in an image, comparing and smoothing these detection using pre-available description text.

	In the same context Farhadi et al. \cite{farhadi2010every} proposed a system that can obtain a score by linking an image to a sentence. In \cite{kuznetsova2012collective}, authors have proposed a data driven approach for description generation by giving a query image and retrieving existing human composed phrases which describe similar images. Combining those phrases they generate a description for the given image. Verma and Jawahar \cite{verma2014im2text} have proposed a system that achieve task like given an image, generating a textual description and vise versa, where both approaches are retrieval based. In the same line Zhu et al. \cite{zhu2015aligning} provides an approach to align visual contents of a movie release to their corresponding book by providing a description of the visuals. Image description is generated by creating visual dependency representation of natural images in \cite{elliott2013image}. Natural language description generation is also done for video for their retrieval purpose in \cite{khan2017generating}, in which they capture relations between keywords associated with videos.	Evaluation of machine translation with human generated description is also necessary. For that purpose several metrics for example BLEU \cite{papineni2002bleu}, ROUGE \cite{lin2004rouge}, METEOR \cite{denkowski2011meteor} etc,  have been proposed. In \cite{elliott2014comparing}, authors provided a correlation between automatic metrics and human judgments, using previously mentioned metrics and their variants.  
	Various approaches have been developed to connect two domains, natural language processing and document image analysis. For example in \cite{chen2015integrating} authors have gained performance by integrating features from linguistic analysis, image text recognition and image layout analysis. Looking at the existing literature we see that even though description synthesis from natural images has become common, the same from document images is still not there, and thus we propose SUGAMAN to bridge the gap. SUGAMAN is an extension of our earlier work \cite{shreya2018}. The key differences between SUGAMAN and \cite{shreya2018} are: (1) A feature based automatic room annotation learning method is proposed, as compared to OCR based method, (2) An improved proximity based sentence model is proposed in SUGAMAN, in stead of a template based model, (3) introduction of room to room navigational information for obstacle free movement. 
    
	\begin{figure*}[t]
		\centering
		\includegraphics[scale=.60]{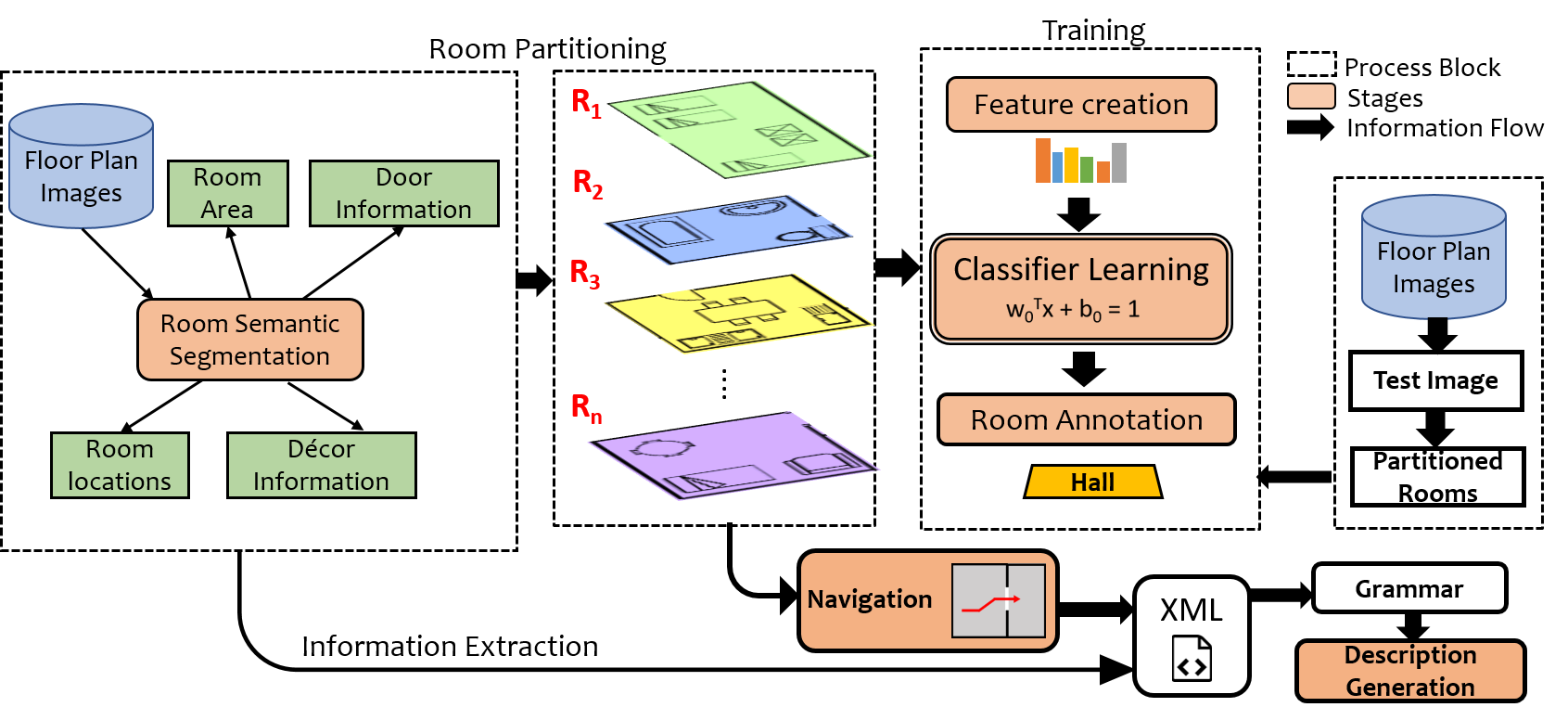}
		\caption{Block diagram depicting various modules and work-flow within SUGAMAN.}
		\label{fig:figure2}
	\end{figure*}

\section{System Overview}
\label{briefdescription}

Figure \ref{fig:figure2} shows a block diagram depicting various modules and work-flow within SUGAMAN. The whole system is divided into two stages (i) room annotation learning and (ii) description synthesis. At first, room semantic information is extracted by room segmentation process, which gives all the required information about an input floor plan image. For example, individual room area, door information which gives room neighborhood information, room locations (room coordinates).  With those room locations, the floor plan image is partitioned into individual room image and taken as sample for room annotations learning. Decor characterization is applied over these room images and decors present in a room are labeled. We have proposed a Local Orientation and Frequency Descriptor (LOFD) feature, which is extracted from these room samples for automatic room annotation. A classifier is trained using LOFD feature matrix of room samples by assigning class labels to them. After this a new input image is taken as a test sample, features are extracted and room annotations are identified for it using previously trained model. An XML file is generated using the semantic information extracted by room segmentation and room classification. By parsing this XML file, textual description is generated. SUGAMAN also gives navigation path within the entire floor plan, starting from the entry door to the building. All such information about floor plan and navigation are fed to the proposed grammar model. The first stage of the proposed description synthesis method deals with \textit{``what to say''} about the floor plan and the second stage will deal with \textit{''how to say it''}. For ease of understanding, in rest of the paper, we demonstrate all our analysis on the input image shown in Fig. \ref{fig:problem}. Later, in the experimental results, we also show the results on other floor plan images. Next we discuss about the dataset used for experimentation.

\section{Floor Plan Dataset}
\label{sec:dataset}
For room segmentation, symbol spotting, retrieval in floor plan images three public dataset were proposed. They are: (i) Systems Evaluation SYnthetic Documents (SESYD) \cite{delalandre2010generation} (ii)  Computer Vision Center Floor Plan (CVC-FP) \cite{de2015cvc} and (iii) Repository Of BuildIng plaNs (ROBIN) \cite{DANIEL2017}. SESYD has ten classes of floor plans, with $100$ samples/class. On the other hand, CVC-FP has $122$ scanned floor plan documents divided into four categories based on the origin and style. In ROBIN there are three broad categories, which are different from each other in terms of the number and type of rooms present in a floor plan. The three categories are (i) $3$ room, (iv) $4$ room, and $5$ room floor plans. Each category is further classified into $10$ sub-categories depending upon the global layout of the floor plan. ROBIN helps in better visualization of the floor plans and aids in efficient capturing of various high-level features while fine-grained retrieval. Since ROBIN has significant number of floor plans, as well as intra-class similarity and inter-class dissimilarity, it is suitable in our case. However, in ROBIN there is no textual description available for a given floor plan. For our purpose we further augmented ROBIN dataset by introducing textual description for each floor plan image. 
 \begin{figure}[!b]
	%	\centering
	\begin{center}
		\includegraphics[scale=.45]{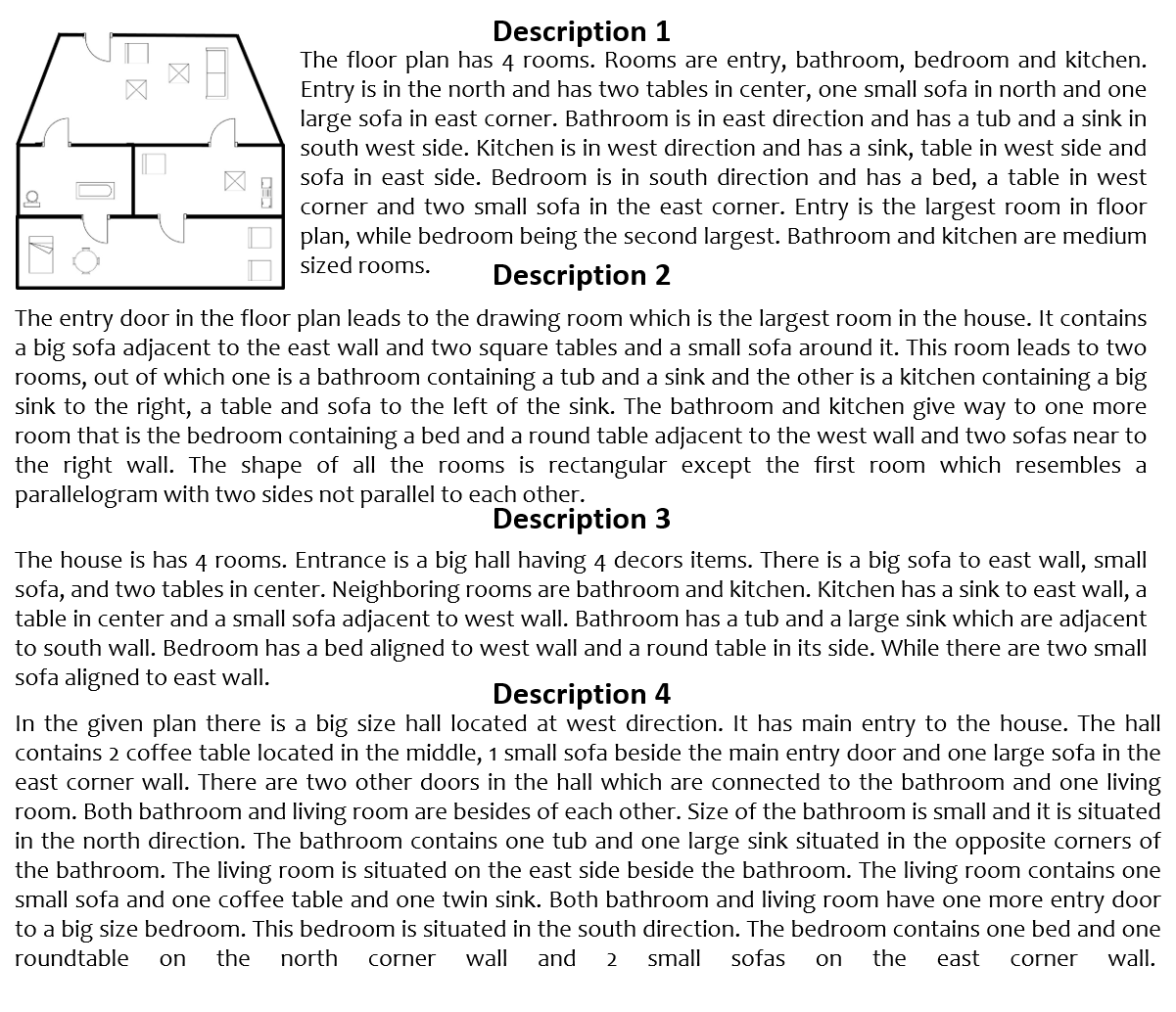}
		\caption{Example of a sample floor plan image from ROBIN dataset and the annotation collected for the same to synthesize A-ROBIN}
		\label{fig:dataset}
		\end{center}
	\end{figure}
\subsection{A-ROBIN}
In order to understand the floor plan images better, and produce automated textual description narrating them we propose a dataset \textit{Augmented} ROBIN (A-ROBIN). In A-ROBIN there are four human written descriptions for each image in ROBIN dataset. In there literature, there are a few image datasets for example Flicker8k \cite{hodosh2013framing}, COCO dataset \cite{chen2015microsoft}, which has associated descriptions of the image samples. However, these datasets describes the natural images. In the context of document images, dataset consisting description of floor plan images was lacking. Hence we require a novel dataset which contains annotations and descriptions for document images. 
The descriptions for a floor plan images were collected from the volunteers in the following manner. Each volunteer was supplied with a set of $10$ images in a Google form and asked to describe them in their own words based on a set of instructions. Each form was given to $4$ volunteers, so that each image has $4$ set of descriptions. These descriptions focus on the rooms and their decor content in the floor plan images. Also they focus on their relative positioning in respective rooms, and relative position of each room within the floor plan image. These descriptions vary in the sequence of the information given, the details of information provided, the sentence conjunctions and the vocabulary used for components in floor plan image. Figure \ref{fig:dataset} shows one of the floor plan and its corresponding descriptions. It can be observed that different descriptions of the same image vary in the amount of information provided, the sequence of describing each room, the names for each decor item could be different for different user. Also, a room may have variations in its name for different user. The length of the descriptions provided for each image is also varied. After the descriptions were collected, each set of $4$ description were tagged to their respective images using image identifiers. 
	\begin{figure}[!b]
	%	\centering[t]
	\begin{center}
		\includegraphics[scale=.55]{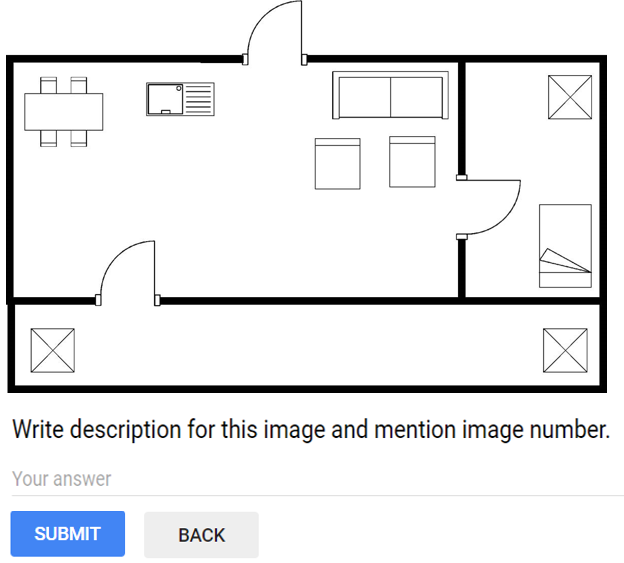}
		\caption{User Interface used for data collection}
		\label{fig:ui}
		\end{center}
	\end{figure}
Following were the instructions given to users:
\begin{itemize}
\item Write information about the whole plan (number of rooms, name of the rooms).
\item Write information of individual rooms- name, decor contained, relative size (small, medium, large, etc).
\item Write information about the relative positioning of decors (north, east, west, south, aligned with wall, adjacent with something, etc).
\item Write information about the relative position of each rooms ( north, east, west, south, adjacent with bathroom, north of kitchen, etc).
\item The description can have this information but they are not restricted to this. You can choose your own words and language. 
\end{itemize}
Figure. \ref{fig:ui} shows the user interface used to collect the descriptions. For $510$ images in ROBIN datasets, there were $2040$ descriptions collected in total. The dataset is tokenized and pre-processed for further processing. In our experiments we have compared machine generated description with these descriptions and an analysis regarding closeness of machine translations with human written descriptions is given. Next, we describe the steps in our framework in details.  

	\begin{figure}[t]
	%	\centering
	\begin{center}
		\includegraphics[scale=.35]{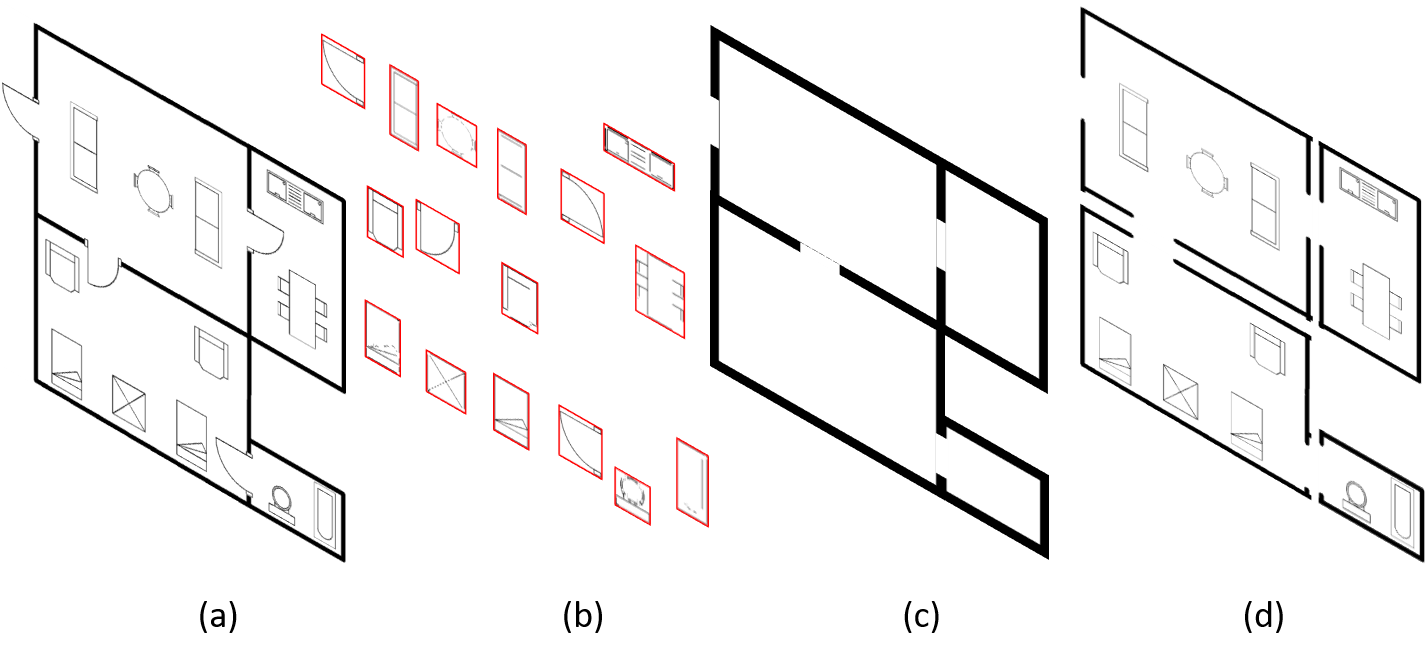}
		\caption{Room segmentation and room partitioning process}
		\label{fig:segmentation}
		\end{center}
	\end{figure}
	
\section{Semantic Segmentation and Room Classification}
\label{sec:RC}
 In all the previous approaches available in the literature, rooms have been classified by recognizing the textual label present in the floor plan image using Optical Character Recognition (OCR) techniques. Room classification in floor plans is not done by extracting salient feature from it. Room classification on the basis of their functionality is very useful in building information modelling (BIM). When a person enters a room in a house, he or she tells the functionality (class) of the room by looking at the decor items present inside the room. This inspired us to propose a unique feature for room classification. We have proposed a new feature called Local Orientation and Frequency Descriptor (LOFD), which represents the frequency of decors present in a room and their normalized distance from the center of the room. We proposed room classification approach as a $5$ class classification problem, which annotates each room in a floor plan into one of the $5$ classes namely, BEDROOM (label-$1$), BATHROOM (label-$2$), ENTRY (label-$3$), KITCHEN (label-$4$), HALL (label-$5$). The following subsections describes the details of room label learning and classification.
   \begin{figure}[!b]
     \centering
     \includegraphics[scale=.50]{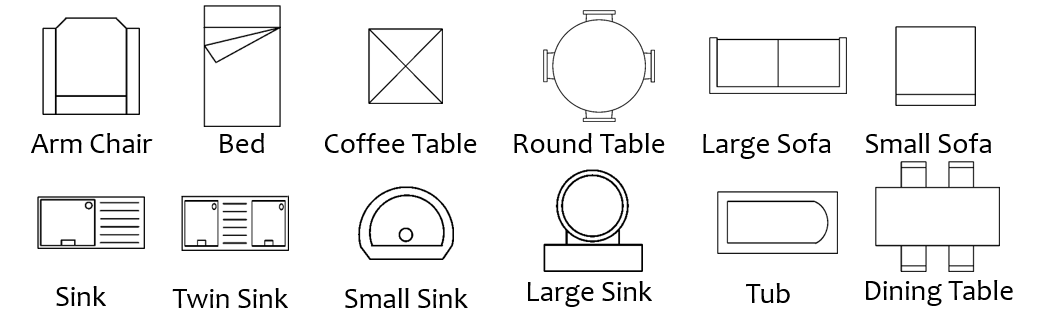}
     \caption{Twelve classes of Decor models used in the experiments.}
     \label{fig:decor}
 \end{figure}
 
\subsection{Room Segmentation}
	\label{sec:segment}
	We have adopted the technique proposed in \cite{divya2016} for the identification of rooms. Walls are detected by performing morphological closing on the input floor plan image $\mathscr{I}$ (see Fig. \ref{fig:segmentation}(c)). To delineate room boundaries, we detect doors using scale invariant features and close the gaps in wall image corresponding to the door locations. To obtain the rooms, we identified the connected components in the wall image by applying flood fill technique. The obtained connected components are the required rooms and their locations are obtained. Also, we calculate the areas of the respective rooms (polygon area), converted them into square feet (taking $100$ pixels= $1$ feet) and store all the information obtained, that is neighborhood, room area, room location coordinates, in a separate data structure. 
\subsection{Floor Plan Partitioning}
\label{sec:crop}
 A floor plan image is partitioned into rooms using the room coordinates extracted from the previous steps as shown in Fig. \ref{fig:segmentation}(d). These individual room images are the samples taken for training the room annotations. We have applied decor characterization in further stages on each of these individual room images to extract the features.

 \begin{figure*}[t]
     \centering
     \includegraphics[scale=.55]{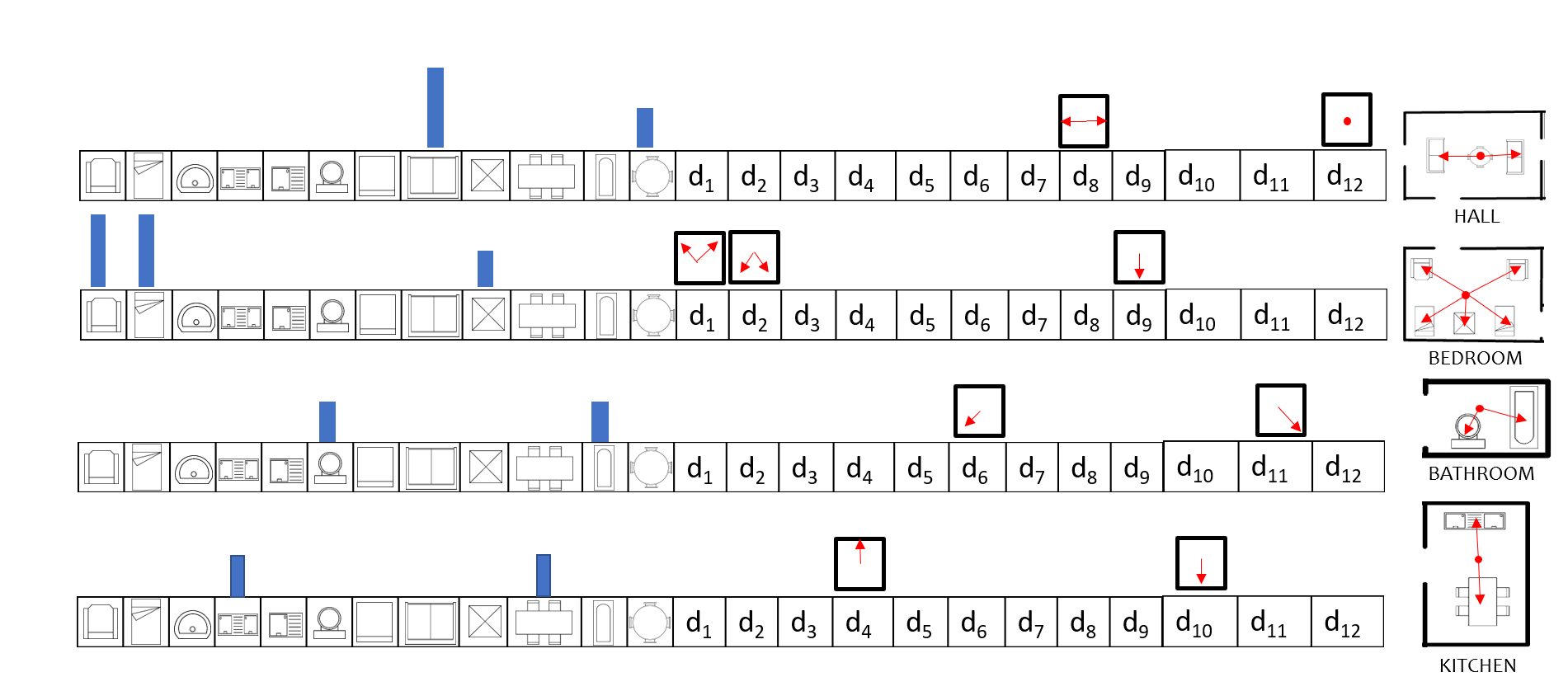}
     \caption{Example of LOFD descriptor for four sample room images}
     \label{fig:feature}
 \end{figure*}
 
 \subsection{Decor Classification}
 \label{sec:decor}
In this section we describe the procedure employed for decor characterization and their classification. Figure \ref{fig:decor} shows the $12$ decor symbols used in the dataset \cite{DANIEL2017}. We have improved the technique of decor characterization proposed in \cite{divya2016} by applying sequence of morphological operations. The technique in \cite{divya2016} uses a normalized area ratio of largest three components of a decor symbol for classification and characterization of decors. We have improved the technique by first collecting $10$ different signatures for each symbol, taking an mean over them (symbols with different orientations) and stored them in a signature library. During classification, we first pre-process the symbol by applying a sequence of morphological operations (erosion and dilation), so that the symbol do not have broken lines. Then we applied blob detection over the image an cropped each decor symbol for signature comparison. Now we compare the test image's signature with the signature stored in library and closest one is classified in its respective category. This modification in the technique greatly improved the classification accuracy for some symbols. Figure. \ref{fig:segmentation} (b) depicts the detection of symbols in the floor plan input image Fig. \ref{fig:segmentation} (a) with bounding boxes. These decors are classified in their respective categories shown in Fig. \ref{fig:decor}.

 \subsection{Local Orientation and Frequency descriptor (LOFD)}
 \label{sec:feature}
Once the decors inside a room are recognized, we compute the feature to classify a room. For room classification, we proposed a new feature named Local Orientation and Frequency descriptor (LOFD). The LOFD feature is required since no other feature descriptor for example SIFT \cite{lowe2004distinctive}, LBP \cite{ojala2002multiresolution} or SURF\cite{bay2006surf} could capture the room images clearly. The LOFD feature is a $1 \times 24$ vector containing the decor information of a room sample and their locally aggregated spatial information. Figure. \ref{fig:feature} shows the LOFD feature matrix for the sample floor plan image. In LOFD, we have aggregated local information of room image in a vector form. LOFD is compact representation of frequency of the decor items and normalized distance of their centers from center of the room. The first $12$ cells of the vectors are occupied by the $12$ decor items from $D=\{D_1,D_2,...D_{12}\}$ and next $12$ cells are occupied by their normalized distances as $\mathscr{D}=\{d_1,d_2,...d_{12}\}$ where $d_n$ is the distance of each decor item from the center of the room. Here, $d_n$ is calculated as:
\begin{equation}
    d_n=\frac{\sum_{i=1}^{k} dist(\mathcal{R}_c,\mathcal{{D}}_c)}{max(\mathscr{D}_n)}
\end{equation}
Here, $d_n$ is the normalized distance for each decor item, $i$ is the count of each decor which may go up to $k$, which is the maximum number of a that decor item in the room, $dist$ is the Manhattan distance between room center $\mathcal{R}_c$ and decor center $\mathcal{{D}}_c$, $max(\mathscr{D}_n)$ is the maximum of all the distances obtained for all the decors to normalize the distance value. Hence LOFD feature distinguishes each room uniquely by the frequency of each decor item and their spatial location in the room. 

 \begin{figure}[t]
    \centering
    \includegraphics[scale=0.40]{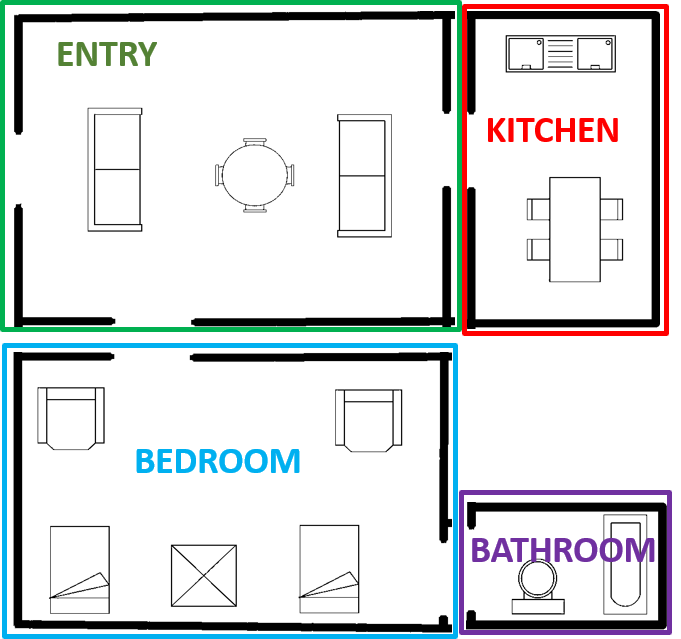}
    \caption{Output of room classification framework using LOFD }
    \label{fig:roomresult}
\end{figure}

Since there are $12$ decor models as shown in Fig. \ref{fig:decor}, first $12$  elements of LOFD represents count of one decor item. However it is not necessary for any room to have all types of decor present, therefore LOFD is sparse in nature. In Fig. \ref{fig:feature}, depicts the room image followed by the corresponding LOFD feature vector. The colored bar over each cell represents the frequency count for each decor item, while arrows in red represents their relative spatial location in the room. In the next section training of the classifier using LOFD feature for room classification is explained.

 \subsection{Room annotations Learning and Room classification}
 \label{sec:label}
 
 Room annotations for training samples (divided the $1355$ room images into $70\%$ and $30\%$ for training and testing respectively) are learned by LOFD feature and classifier. For training purpose, we have manually annotated the room samples, and used those annotations during training. Extensive experiments were performed using various classifiers and the best classifier in term of highest training accuracy is taken for testing the model. For testing purpose, an image from test set is taken and class labels are evaluated accordingly for the room samples of that floor plan image. For each new test floor plan image, feature vector is evaluated for every room. Therefore dimension of feature matrix for a test floor plan image will be $N_r \times 24$ where $N_r$ is the number of rooms in the floor plan. Trained classifier is used for this feature matrix and output class labels are evaluated. Figure \ref{fig:roomresult} depicts the annotations obtained for each room in a floor plan image, where different colors signifies different rooms and their respective annotations. Thus for a given floor plan we obtain room names and the decors within the rooms. 
 \begin{figure}[t]
    \centering
        \includegraphics[scale=.27]{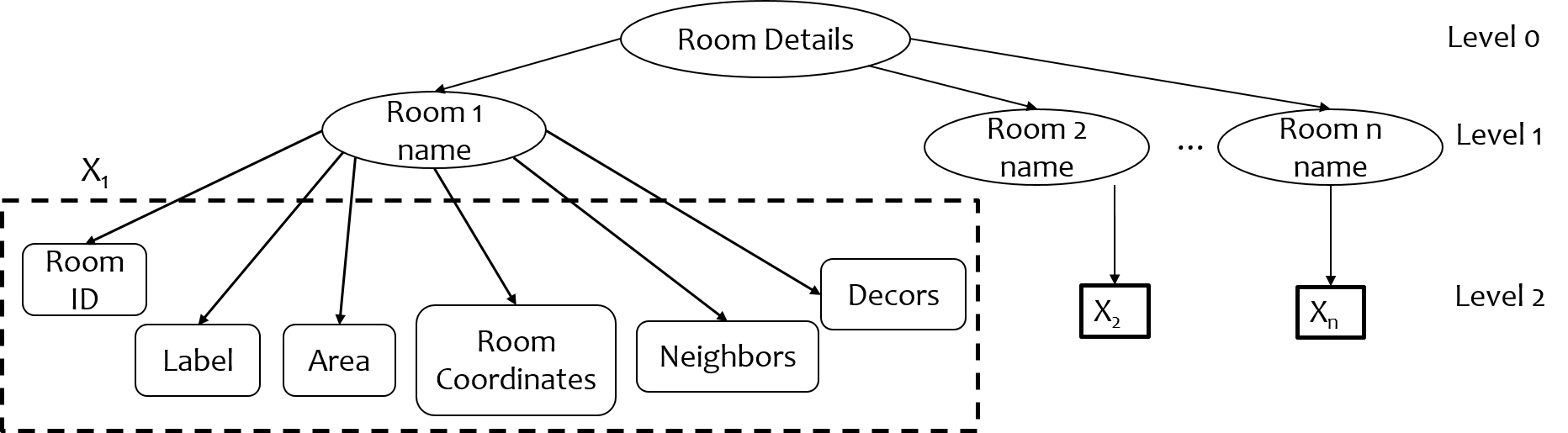}
    \caption{Structure of the XML file generated for description synthesis }
    \label{fig:xml}
\end{figure}

\section{Description Synthesis}
\label{sec:description}
Rooms are classified and their annotations are learned in Sec.\ref{sec:label}. Information extracted from room segmentation are combined and used for generating the description of the floor plan image. Information related to individual rooms are combined and stored in an XML file, which is parsed to generate description of the floor plan.  

\begin{figure}[!b]
    \centering
    \includegraphics[scale=0.40]{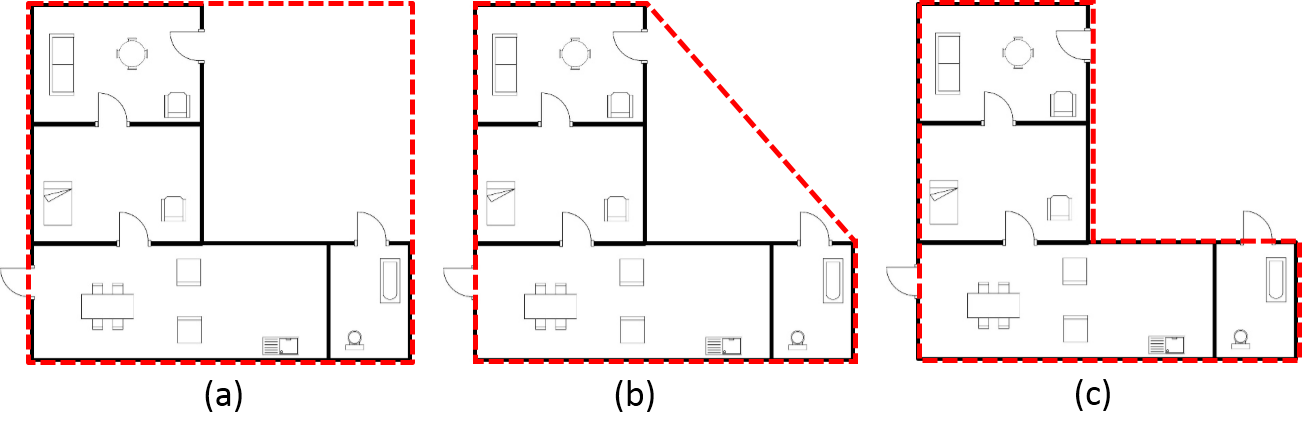}
    \caption{Boundary tracing example for different values of t}
    \label{fig:boundary}
\end{figure}

\subsection{XML File generation}
\label{sec:xml}
An XML file has many benefit in terms of cross platform portability, ease of understanding by novices, and extendability. We have created an XML file by combining the semantic information extracted from room segmentation and room annotations learned in previous steps. As shown in Fig. \ref{fig:xml}, the tree like structure of XML file contains ``Room details'' as root node at level $0$, ``Room names'' as nodes at level $1$, and information of rooms as nodes at level $2$ (leaf nodes), which are Room ID, Room annotations, Room area, Room Coordinates, Room neighbors and Room Decors. Apart from room annotations, a room ID is given to each room since room annotations can be same for two rooms. 
to generate a description.

\subsection{Coordinates systems}
\label{sec:coords}
For defining the positions of rooms and decors present in the floor plan, we have defined two coordinate systems. The global coordinate system is to identify the global location of rooms with respect to the entire document. Local coordinate system is to define relative position of decors with respect to each room. 

\begin{figure}[t]
    \centering
    \includegraphics[scale=0.30]{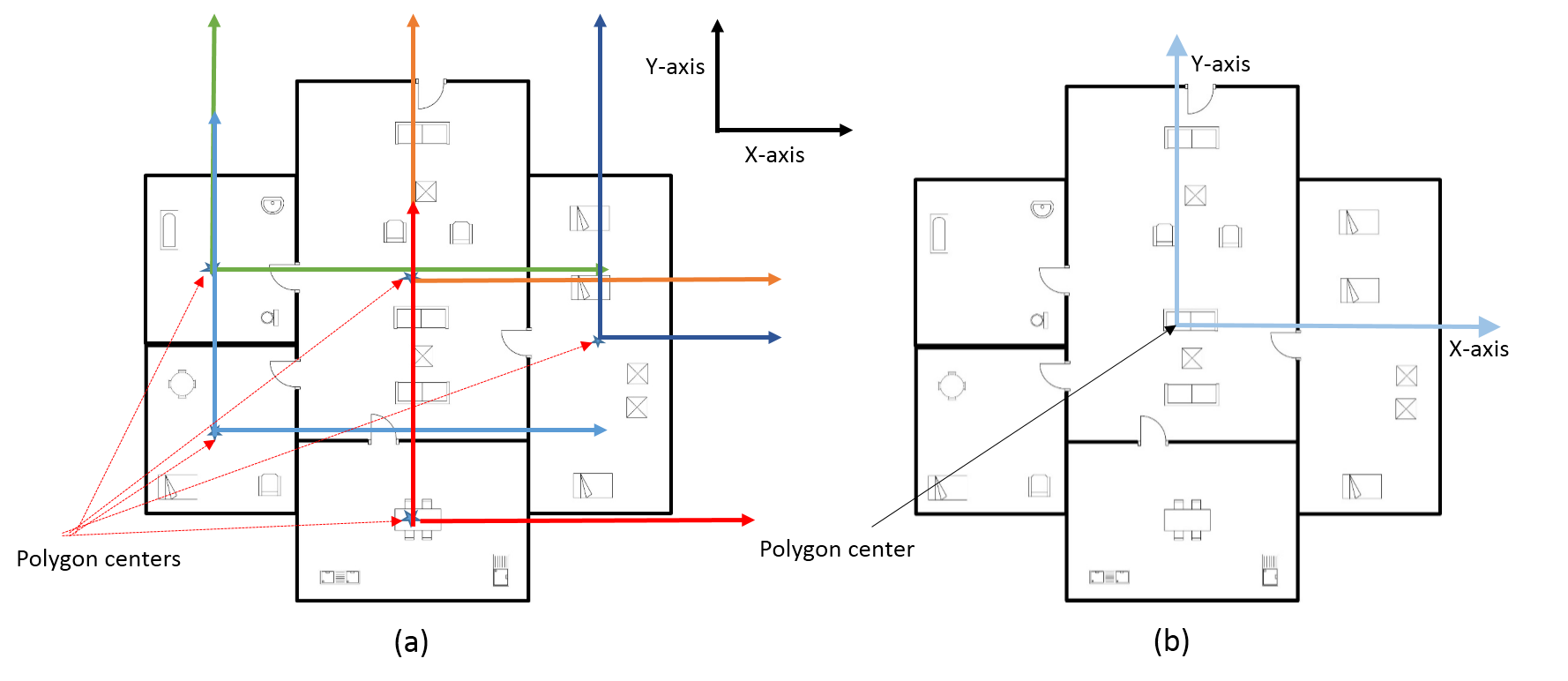}
    \caption{Global and local Coordinate systems}
    \label{fig:Global}
\end{figure}

\subsubsection{Boundary Tracing}
\label{sub:bound}
The origin of the global coordinate system is the center of the polygon, which makes the boundary of the floor plan. To identify the center of the floor plan, its boundary is traced. In order to trace the boundary, coordinates which compose individual rooms are plotted and an outer boundary is tracked which encloses all the outer points, since these points collectively makes the floor plan image. However, by tuning the value of shrinkage factor $t$ between $0$ and $1$ we can switch between a convex hull of those points and a more close knit boundary. Shrinkage factor defines how closely the hull envelops the boundary points. For example, in Fig. \ref{fig:boundary} (a), the boundary traced is a convex hull for the floor plan image for the shrinking factor value $t=0$, Fig. \ref{fig:boundary} (b) is the traced boundary for shrinking factor value $t=0.5$ and Fig. \ref{fig:boundary} (c) is the close knit boundary for $t=0.8$. Hence by tuning the shrinking factor value we can obtain a close knit boundary for the floor plan image. 

\subsubsection{Global and local coordinate systems}
A global coordinate system defines the global position of all the rooms in a floor plan image (see Fig. \ref{fig:Global} (b)). From the traced boundary obtained in the previous step, we calculate the origin of the global coordinate system. Equation \ref{eq:area} and \ref{eq:center} lists the governing equations.

\begin{eqnarray}
\label{eq:area}
a_i&=&x_i y_{i+1} - x_{i+1} y_i \\ \nonumber
A&=&\frac{1}{2}\sum_{1}^{n}a_i 
\end{eqnarray}

Where, $a_i$ in Eq. \ref{eq:area} is twice the signed area of the elementary triangle formed by $(x_i,y_i)$ and $(x_{i+1},y_{i+1})$ and the origin. $A$ in Eq. \ref{eq:area} is the area of the polygon.

\begin{eqnarray}
\label{eq:center}
x_c&=&\frac{1}{6A}\sum_{1}^{n}a_i(x_i+ x_{i+1}) \\ \nonumber
y_c&=&\frac{1}{6A}\sum_{1}^{n}a_i(y_i+ y_{i+1})  
\end{eqnarray}

In Eqn. \ref{eq:center}, $(x_c,y_c)$ is the center of the polygon. The local coordinate system (see Fig. \ref{fig:Global} (a)) identifies the relative positions of all decors with respect to each room. Center of each room, for a local coordinate system is computed using Eq. \ref{eq:area} and \ref{eq:center}.
%\begin{figure}
 %   \centering
  %  \includegraphics[scale=0.40]{localcoord.png}
   % \caption{Local Coordinate system}
    %\label{fig:local}
%\end{figure}
\subsection{Binning}
\label{sec:bin}
\begin{figure}[t]
    \centering
    \includegraphics[scale=0.25]{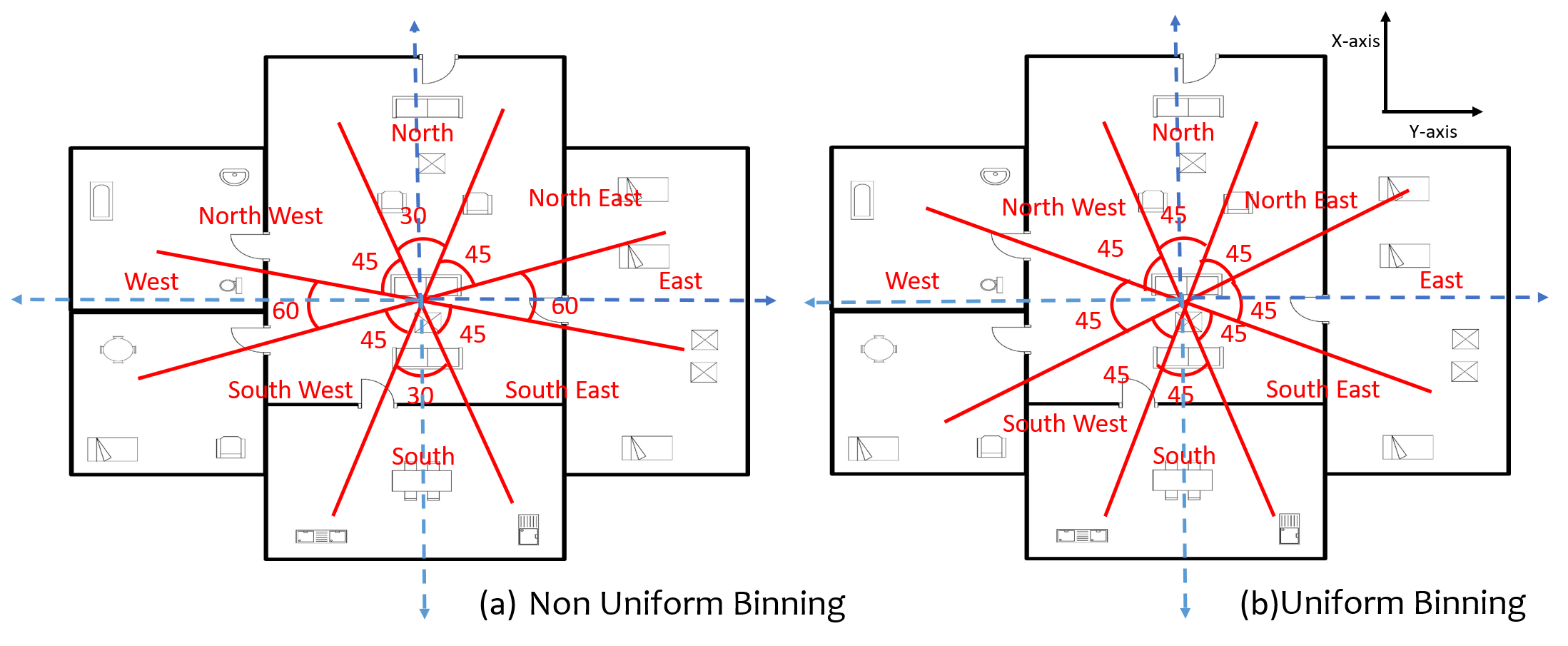}
    \caption{Non uniform and uniform binning}
    \label{fig:bin}
\end{figure}

We have performed global and local binning or radial partitioning of the floor plan (see Fig. \ref{fig:bin}). The non uniform binning angles were empirically determined. For identifying the direction of a decor, the center of the surrounding bounding box is taken as the reference point. While for the rooms, their respective centers, obtained in the previous steps is taken as reference point. As shown in the Fig. \ref{fig:bin}(a), the entire coordinate system is divided into $8$ directions, north, north-east, east, south-east, south, south-west, west, north-west, in the clockwise direction. The binning depicted in Fig. \ref{fig:bin}(a) is a non uniform binning, while in Fig. \ref{fig:bin}(b) is a uniform binning.

\begin{figure}[!b]
    \centering
    \includegraphics[scale=0.30]{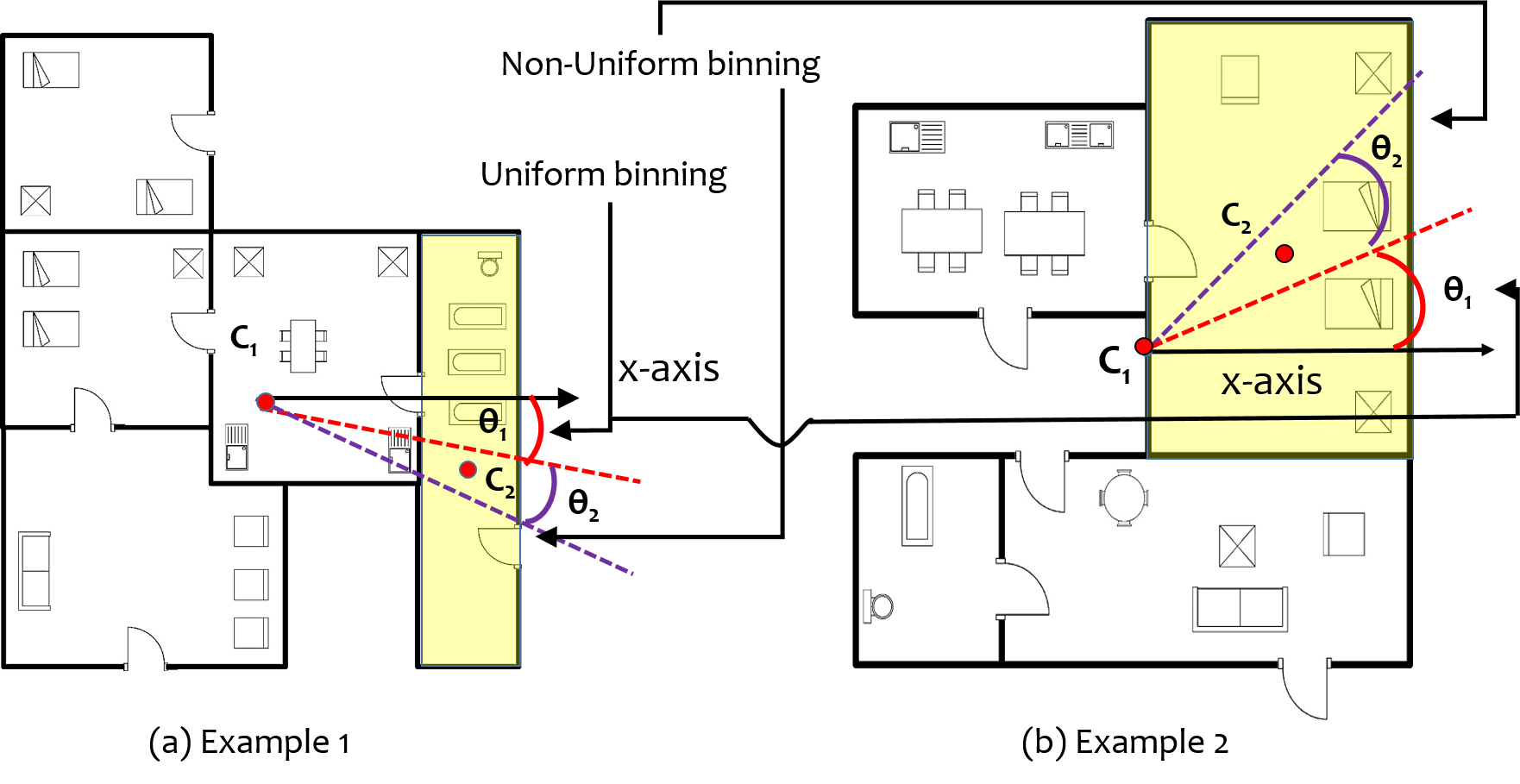}
    \caption{Illustration for the rationale of non-uniform binning.}
    \label{fig:binex}
\end{figure}

The rationale behind non-uniform binning is to provide a more realistic direction information for rooms and decors. The idea of taking the direction from the center of the surrounding polygon may misguide the framework about the actual position of a room. E.g., if a room location in the west direction and stretched towards north, its center will lie in north west direction even if the room is in west. In order to avoid these kind of ambiguities, binning is done non uniformly and the angles are empirically taken. Figure \ref{fig:binex} highlights examples for the above rationale. The highlighted room (Fig: \ref{fig:binex} (a)) is more toward east direction, however it is also extended towards south. With non uniform binning we try to increase the span of east direction, shown purple line and arc where $(\theta_1+\theta_2)$ is the angle of non uniform binning. While red line and arc shows the span of uniform binning which makes the room fall in south east direction and create ambiguity. Here $\theta_1$ is the angle of uniform binning, $C_1$ and $C_2$ are the centers for floor plan and room respectively. 

\subsection{Navigation}
Navigation in the indoor environment by avoiding the obstacles is an integral part of SUGAMAN. We have proposed a grammar based model that yields navigational directives to navigate through the house for a natural movement, from one door to the other door of each room. The algorithm is divided into two parts, first we create a data structure, which stores the room labels along with their respective doors and their corresponding index. The room information and the door coordinates are obtained from semantic segmentation in the earlier stages (see Sec. \ref{sec:segment}). If a door is shared between two rooms then that door will be present in both room's door structure and the index will represent the door's identity. Next, we identify the entry room and the corresponding door, and generate a Depth First Search (DFS) ordering of the region adjacency graph of the floor plan taking the entry room as the start node. After that, a path to the next room is generated avoiding obstacles, by checking the visibility from first door to the other. We also create a door based adjacency matrix ($AM_D$), which stores the shared doors between rooms. 
 \subsubsection{Creating door structure}
The room coordinates, room labels and door coordinates are obtained in the semantic segmentation. After that an index $i_d$ is assigned to each door. We have checked whether a given door $i_d$ belongs to a particular room or not. We have performed an inside-outside test between the bounding polygons of the doors and the rooms to achieve the belongingness. The door structure contains each room with its corresponding doors having marked with their index $i_d$. As shown in Fig. \ref{fig:nav}(b), room and door information is stored in a door structure. 

\subsubsection{Path Finding}
DFS search is performed over the region adjacency graph of the floor plan image taking the entry room as starting node. The door connected to the outer wall of the entry room is considered as the entry door and stored in the door structure. Here, entry door for the house is detected by the algorithm discussed in \cite{shreya2018}. Algorithm. 1 describes the process of room to room navigation by obstacle avoidance. The route in each room is stored in the form of coordinates of movement and included in the description for narration of the path. Algorithm 1 traverse the rooms starting from the first room in the DFS graph, by checking if there is a door shared between them. This is checked by door based adjacency matrix ($AM_D$). If they do not share a door, the algorithm backtrack and explore other rooms. Also, it determines the route across the rooms for navigation. 
In Alg. 1, line 2 declares the flag, if the algorithm has to enter into backtracking. Line 1 describes the loop which traverse room to room finding the path. Line 5, algorithm checks if there is a shared door between the current room and the next room and continues traversal between rooms if there is a shared door. Line 6 directs the algorithm to further processing if backtracking is not required. Line 7 to 10, detects the coordinates of bounding box of decor items and centroid of doors of current room and include in a vertex list. In 11, doors are removed from room image because they are not required for avoiding obstacles. Line 12 to 15 detect the corner points in the room image using Harris corner detector after detecting the blobs, and include maximum 1000 strongest corners in the vertex list. Line 16 to line 22 describes the construction of adjacency matrix for navigation ($AM_N$). It checks the visibility between every point in the vertex list and include the Euclidean distance between them in $AM_N$ as the weight at $AM_N$($V_L$(j),$V_L$(k)). Visibility between two points is checked by filling the line between equal intervals in those points and checking if there is a black pixel present. If there is a black pixel present, then there must be an obstacle between those two points and hence those points are not visible. $AM_N$($V_L$(j),$V_L$(k)) will have a $0$ in that case. Line 26 and 27 defined the entry ($D_E$) and exit ($D_X$) door for current traversal, where entry door is the entry of current room and exit door is the entry of the next room. Line 29 evaluates a route ($P^i$) for current traversal by applying Disjkstra's shortest path algorithm over $AM_N$ taking $D_E$ and $D_X$ as start and end nodes. Line 31 to 34 defines the backtracking process if there is no shared door found between current room and next room. Algorithm will backtrack in the DFS path and find the navigation path between corresponding rooms. The route for $i^{th}$ room ($P^i$) is a set of coordinates, which contains the start point, end point and intermediate turns which a person have to make for obstacle avoidance. Figure. \ref{fig:nav} describes the entire process for the input image Fig. \ref{fig:nav}(a). The checker box (inset) depicts $AM_D$, where the dark box represents a $0$ and a white box represents a $1$.  

Figure. \ref{fig:nav}(b) shows the door structure created in the previous step and the order of traversal with backtrack step, Fig. \ref{fig:nav}(c) shows the DFS search graph generated over the region adjacency matrix to obtain the order to traverse the each room and Fig. \ref{fig:nav}(d) shows the local coordinate system fitted over every point in a route while traversing through the floor plan, also showing the direction of movement by arrows. Figure. \ref{fig:navex}(a) represents the door to door path generated for navigation, avoiding obstacles in each room for the input image. Figure. \ref{fig:navex} shows some other examples describing the path generated on various floor plan images.  

\begin{figure*}[t]
    \centering
    \includegraphics[scale=0.8]{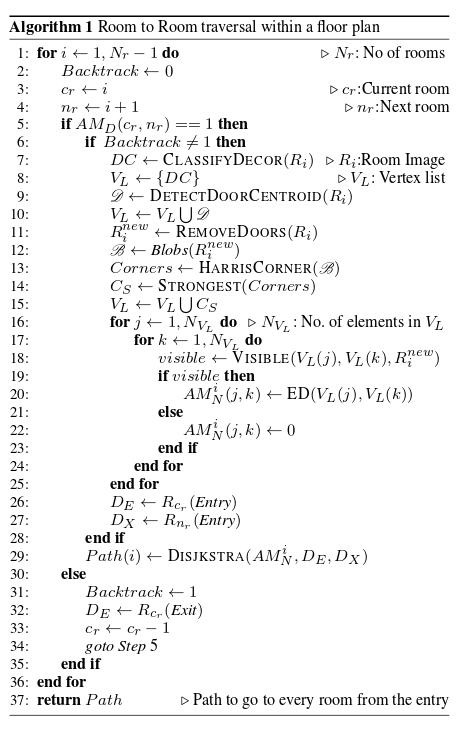}
    \caption{Algorithm for the navigation }
    \label{fig:nav}
\end{figure*}

\begin{figure*}[t]
    \centering
    \includegraphics[scale=0.5]{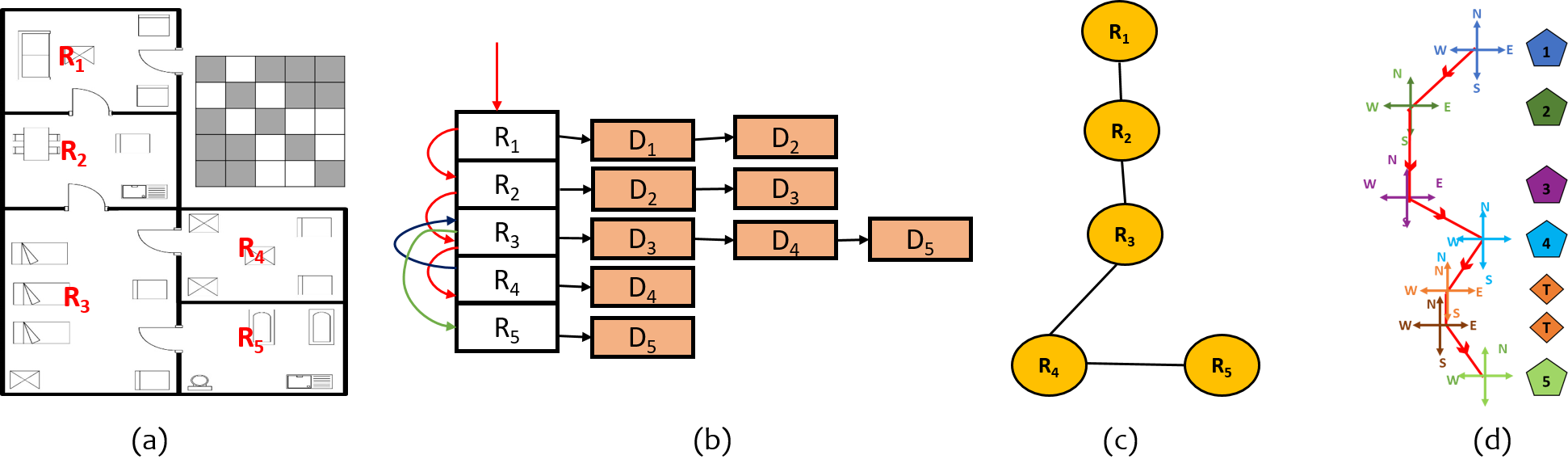}
    \caption{Illustration of path detection process}
    \label{fig:nav}
\end{figure*}

\begin{figure*}[!b]
    \centering
    \includegraphics[scale=0.4]{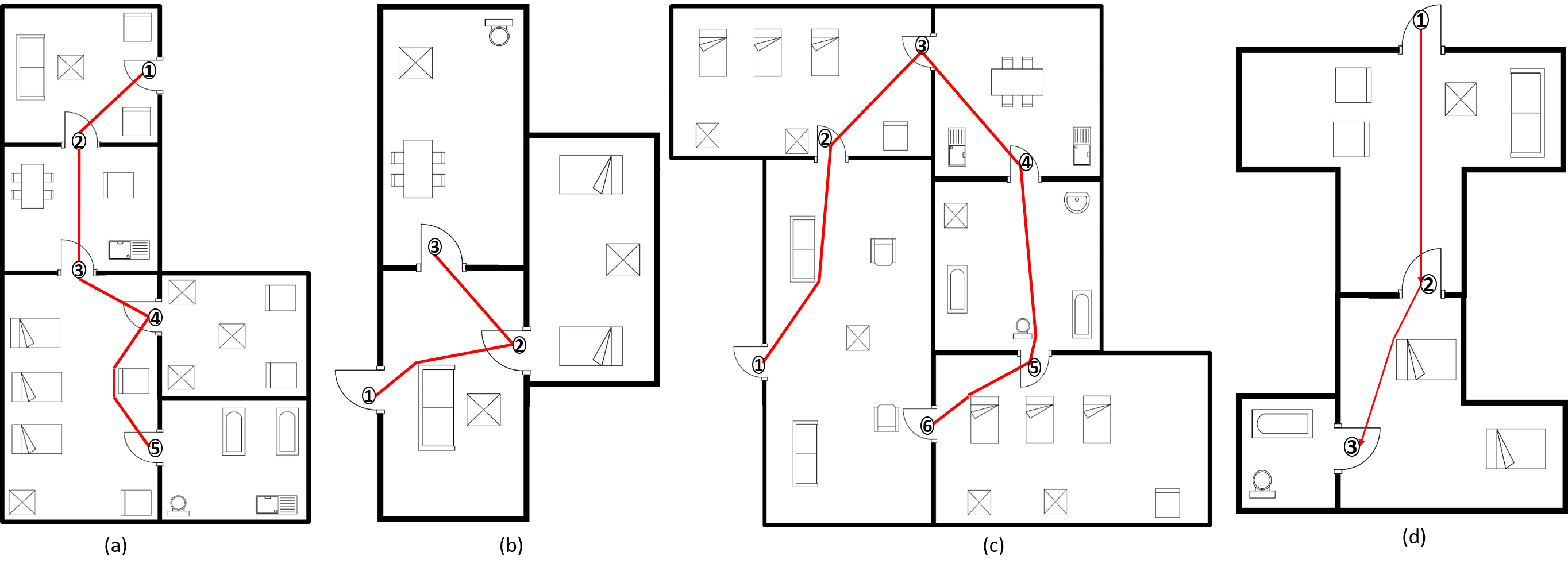}
    \caption{Examples illustrating path detection by avoiding obstacles}
    \label{fig:navex}
\end{figure*}

\subsection{Proximity based sentence model}
Parsing of the XML file yields $5$ types of information for each room, defined in separate sentences; Room name, area, neighbouring rooms, global position and contained decors with their relative position in room. For that purpose we defined, sentence model having $6$ rules, which is based on proximity as shown in Tab. \ref{tab:prox}. The first sentence ($S_1$) of description for every floor plan is a general sentence stating the number of rooms ($N_r$) in the floor plan. In $S_2$, DT is a determiner which takes its value from the set \{a, an\}. Also, $O_i$ is the object which takes its value from level $1$ nodes (Room names), where value $i$ varies from $1$ to $N_r$. In $S_3$, AREA takes its value from the RoomArea tag when XML file is parsed. In $S_4$, $s$ takes its value from the set \{s,$\phi$\}, which is a proximity based value depends upon its previous word. Value $s$ is chosen if the word in proximity (room) is a plural and $\phi$ otherwise. Also, AUX is an auxiliary verb, which takes its value from \{is, are\}, depending upon its proximity word and ${{NR}_j}$ takes its value from Neighbors tag (neighboring rooms), when XML file is parsed. Here, value of j varies from 1 to ${NN}_r$ which is number of neighboring rooms. In $S_5$, LOC is the global position of room which takes its value from the set \{North, North East, East, South East, South, South West, West, North West\} described by binning. In $S_6$, the value of $k$ varies from 1 to $DC$ i.e. decor count. Here, $C$ is the count of individual decor item, $D$ takes its value from the Decor tag in XML file, $s$ takes its value from \{s,$\phi$\} and $DLOC$ is the relative location of decor in the room which takes its value from \{North, North East, East, South East, South, South West, West, North West\} described by binning. 

\begin{table}[!b]
\centering
 \caption{Sentence model based on proximity  }
 \begin{tabular}{||c c ||} 
 \hline
 Sentence  & Rule\\
 %$(\%)$&$(\%)$&$(\%)$\\
 \hline\hline
  $S_1$ & This floor plan has \bm{$N_r$} rooms \\
 \hline
 $S_2$ & There is \textbf{DT} \bm{$O_i$} \\
 \hline
 $S_3$ & It has an area of \textbf{AREA} \\
 \hline
 $S_4$ & Its neighboring room\{s\} \textbf{AUX} \bm${{NR}_j}$  \\
 \hline
 $S_5$ & It is located in the \textbf{LOC}\\
 \hline
 $S_6$ & This room has $\{ \bm{C}\; \bm{D}\{s\}\; at\;the\; \bm{DLOC}\}_{k}$ \\
 \hline
% $S_7$ & Go Straight. \\
% \hline
 $S_7$& $\{ \{ Go \; \bm{N_{step}}\; steps\; in\; \bm{DIR} \;direction\;\}_{N_m}\}_{N_r}$\\
  
 \hline
 $S_8$ & There is a door and a room. $\{S_{9}\}_{if\;dead\; end}$\\
 & $\{S_7\}_{else}$. \\
 \hline
 $S_{9}$ & You have to turn back.\\
 \hline
 %\bottomrule
 \end{tabular}

    \label{tab:prox}
\end{table}

$S_7$ is the sentence narrating the navigation, where $N_{step}$ is the number of steps to be taken. For calculating number of steps we took the euclidean distance between first coordinate and next coordinate in the route and calibrated the distance into steps ($10$ pixels$=1$ step). Also, $DIR$ is the direction in which the person has to move, for which local coordinate system is being fit on every coordinate of the route. It takes its values from the set \{North, North East, East, South East, South, South West, West, North West\}. 
The number of coordinates returned in the route of navigation inside a room, is the number of turns a person will have to make. Here $N_m$ is the number of turns inside a room and $N_r$ is the number of rooms. $S_8$ describes the door and room found after navigating through previous room. If the room has only one door and hence a dead end, the person will turn back and navigate further ($S_{9}$), else he will go straight and explore the other rooms by entering ($S_7$).

%\subsection{Examples of navigation}

\begin{figure}[t]
    \centering
    \includegraphics[scale=0.30]{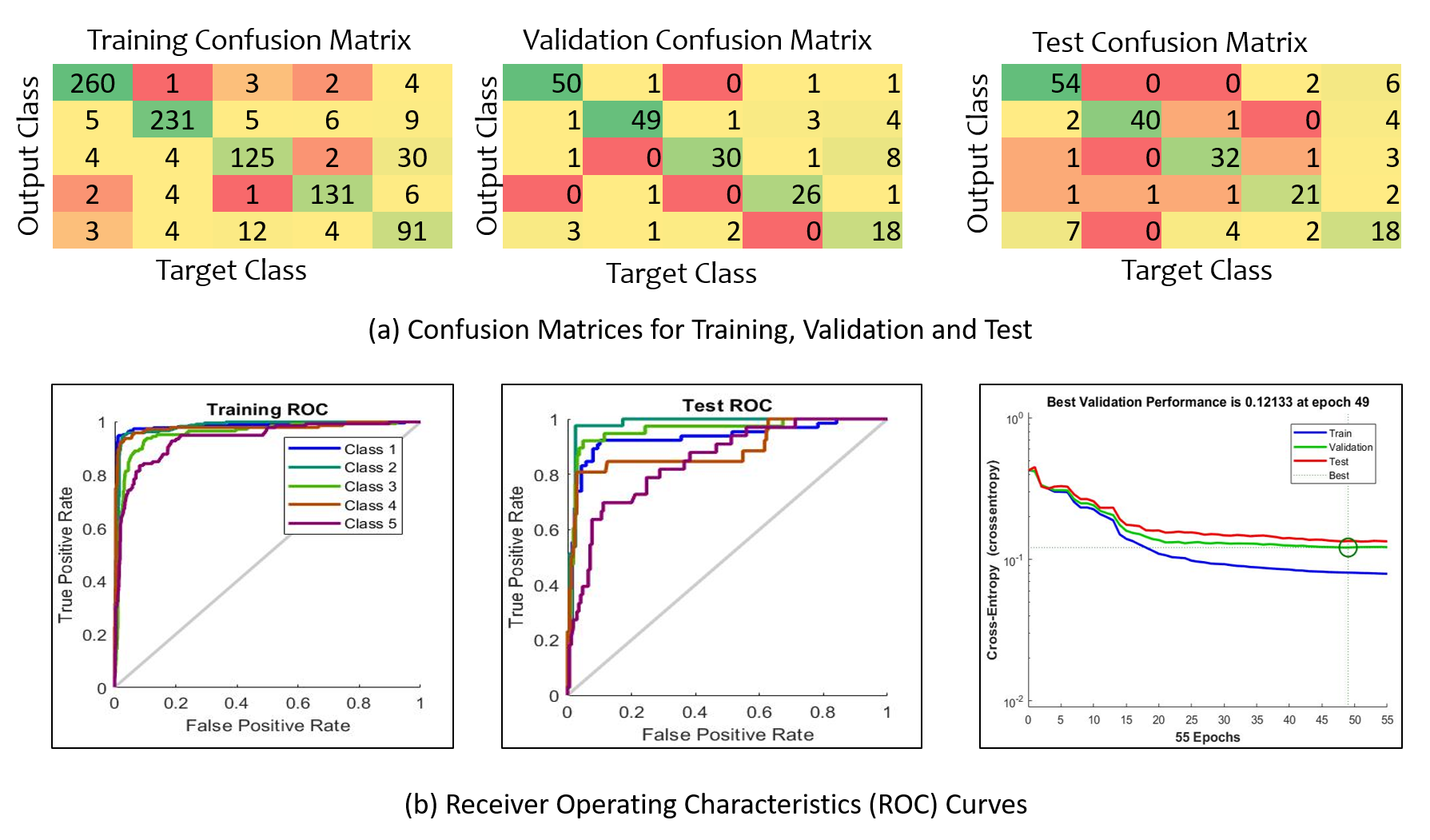}
    \caption{Performance analysis of multi-layered perceptron.}
    \label{fig:nn}
\end{figure}

\section{Analysis of intermediate steps}
\label{sec:exp}
We have performed our experiments on a hardware platform with the following configurations. The system has an Intel core i$7$ ($8^{th}$ generation), with a $1.87$ GHz processor. It has a memory of $8$ GB where, implementation has been done on Matlab $16a$.

\begin{table*}[!b]
%\begin{center}
\centering
\caption{SVM classifier-Results of room annotation learning by support vector machine}
 \begin{tabular}{||c c c c c c ||} 
 \hline
 Variant & Training & Testing & Testing& Training & Testing  \\ [0.5ex] 
 &(R)&(R)&(S)&(R+S)&(R+S)\\
 &  $(\%)$   & $(\%)$ & $(\%)$&$(\%)$&$(\%)$\\
 \hline\hline
 linear svm OVO & $90.1$ & $78$ & $66.04$& $89.1$ &$73.4$  \\ 
 \hline
 Quadratic SVM OVA& $87.8$ & $79$ &  $60$&  $91.6$& $78.4$\\
 \hline 
 Cubic SVM OVA & $87.8$ & $77.11$ & $63$& $91.2$ & $76.6$\\
 \hline
 Medium Gauss SVM OVO & $88.3$ & $76.67$ & $58$ &$90.5$ &$75.3$ \\
 \hline
 Quadratic SVM OVO  & $87.1$ & $77.11$ & $60$ &$90.9$ & $74.2$ \\  
 \hline
 Complex Tree& $88.3$ & $76.44$ &$65.57$ & $88.7$ & $73$ \\
 \hline
 %\bottomrule
\end{tabular}
%\end{center}
    \label{tab:svm}
\end{table*}

\begin{figure*}[t]
\begin{center}
\includegraphics[width=\linewidth]{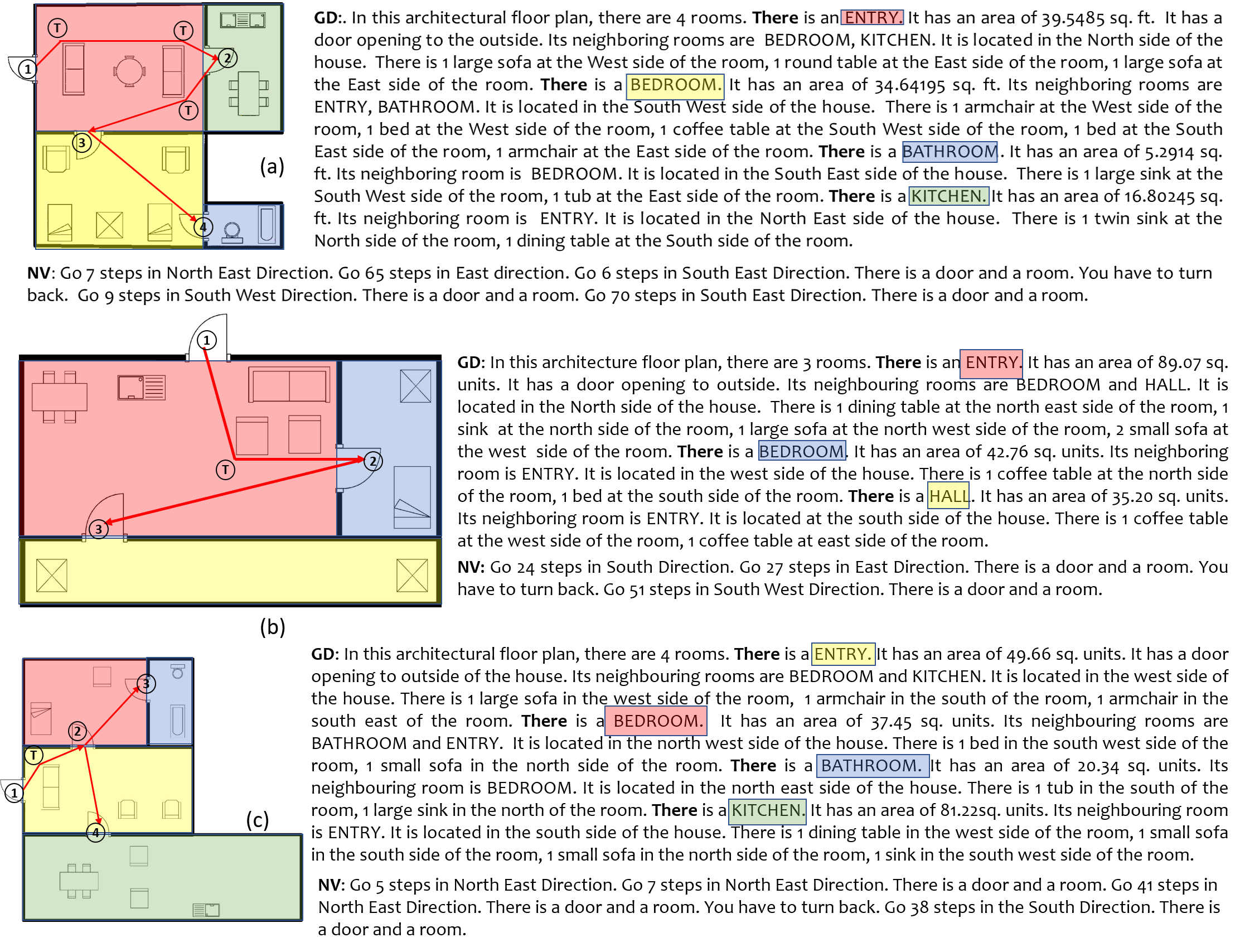}
 \end{center}
 \caption{Generated descriptions for three floor plan images from A-ROBIN dataset.}
 \label{fig:resultdes}
\end{figure*}

\begin{figure*}[!b]
\begin{center}
\includegraphics[scale=0.54]{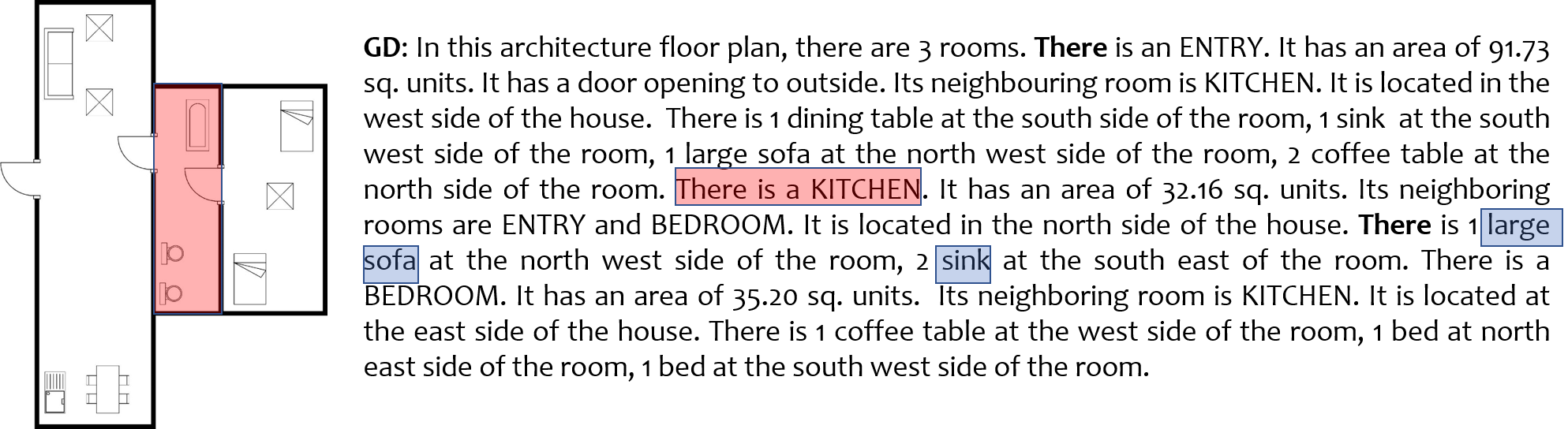}
 \end{center}
 \caption{Generated descriptions for three floor plan images from A-ROBIN dataset.}
 \label{fig:resultdes}
\end{figure*}

\subsection{Room annotation learning and Classification}
For this task a dataset of $1355$ room images divided into $70\%$ and $30\%$ for training and testing respectively. LOFD features (see Sec. \ref{sec:decor} and Sec. \ref{sec:feature}) are used to train a multi-layered perceptron ($1$ hidden layer with $10$ neurons). Performance of neural network is shown in Fig. \ref{fig:nn} using confusion matrix and Receiver Operating Characteristics (ROC) curves. It is clear that ROC curve for class $5$ moves maximum towards false positive axis, because of less number of training samples for class $5$. Also, for class $1$ and $4$ it remains toward true positive axis due to more number of samples in training data. Figure. \ref{fig:nn}(b) (column $3$) shows the performance curve for neural network in which best validation performance is achieved at epoch $49$. The training, testing and validation accuracy obtained by neural network are $88.3\%$, $81.3\%$ and $85.2\%$, respectively. 

\begin{table}[t]
\centering
 \caption{Performance analysis of text generation algorithm using ROUGE score.}
 \begin{tabular}{|c| c| c| c |} 
 \hline
 ROUGE & Average Recall & Average Precision & F score\\[0.5ex]
 \hline\hline
 ROUGE-1 & $0.5061$ & $0.2715$ & $0.3445$ \\
 \hline
 ROUGE-2 & $0.1545$ & $0.5707$ & $0.07616$ \\
 \hline
 ROUGE-3 & $0.0535$ & $0.01093$ & $0.01483$ \\ [1ex]
 \hline
 \end{tabular}
    \label{tab:rouge}
\end{table}

\begin{table}[!b]
\centering
 \caption{Performance analysis of text generation algorithm using BLEU score.}
 \begin{tabular}{|c| c|} 
 \hline
 METRIC & Score  \\
 \hline\hline
 BLEU-1 &  $0.6418$  \\
 \hline
 BLEU-2 & $0.4673$ \\
 \hline
 BLEU-3 & $0.3448$   \\ 
 \hline
 BLEU-4& $0.2103$\\
 \hline
 \end{tabular}
    \label{tab:BLEU}
\end{table}

\begin{table}[t]
\centering
 \caption{Performance analysis of description synthesis using METEOR score.}
 \begin{tabular}{|c|c|c|c|c|} 
 \hline
Average Recall & Average Precision & F1 & F mean & Final Score\\[0.5ex]
 \hline\hline
 $0.555$ & $0.218$ & $0.313$ & $0.450$ & $0.184$ \\ [1ex]
 \hline
 \end{tabular}
    \label{tab:METEOR}
\end{table}

Experimental results on other supervised classifiers are shown in Tab. \ref{tab:svm}. For training the classifiers, as shown in Tab. \ref{tab:svm}, first we divided $1355$ samples from ROBIN(R) dataset in $70\%$ (training) and $30\%$ (testing). Training and testing accuracy are shown in first and second column respectively. We tried testing the sampled from SESYD (denoted as S) dataset from this trained model but testing accuracy statistics (column $3$) are not up to the mark, hence we mixed the samples from both datasets. Taking $500$ samples from SESYD and $1355$ samples of ROBIN  making it a collective dataset of $1855$ images, another models were trained and training and testing are shown in column $4$ and column $5$, respectively. The best performing classifier is linear Support Vector Machine (SVM), one verses one, for ROBIN dataset and quadratic SVM (one verses all) for mixed samples making LOFD a highly accurate feature descriptor for room annotation learning in floor plan images.

\subsection{Description synthesis}
All the Reference Corpus available in the A-ROBIN dataset and the generated descriptions were tokenism using The Penn Treebank tokenizer \cite{marcus1993building} and kept for evaluation purpose. We have compared the machine generated description of the floor plan with human written descriptions in A-ROBIN. The generated description is evaluated by three metrics, Recall-Oriented Understudy for Gisting Evaluation (ROUGE) \cite{lin2004rouge}, Bilingual Evaluation Understudy (BLEU)  \cite{papineni2002bleu} and Metric for Evaluation of Translation with Explicit Ordering (METEOR) \cite{denkowski2011meteor}. The textual description generated by our framework is then compared with the descriptions in A-ROBIN to evaluate their agreement with human written descriptions . Table \ref{tab:rouge} depicts the average recall, average precision and F score for ROUGE-$1$, ROUGE-$2$, ROUGE-$3$. As the value of $n$ in n-gram comparison increasing, the ROUGE precision score decreases, which is also clear from Tab. \ref{tab:rouge}.  Table. \ref{tab:BLEU} depicts the BLEU score and Tab. \ref{tab:METEOR} METEOR score for the description generated, which demonstrates high correlation with human judgments.

\subsubsection{ROUGE}
ROUGE is a set of metrics designed to evaluate the text summaries. The generated summary can be evaluated with a set of reference summaries. In our work, we have compared the generated descriptions with available human written descriptions using n-gram ROUGE by the following equation. 

\begin{equation}
    \frac{\sum_{S\in\{RS\}}{\sum_{gram-n\in S}{Count_{m}(gram-n)}}}{\sum_{S\in\{RS\}}{\sum_{gram-n\in S}{Count(gram-n)}}}
\end{equation}
Where $RS$ stands for reference summaries, $n$ stands for length of the n-gram, $gram-n$, and $Count_{m}(gram-n)$ is the maximum number of n-grams co-occurring in the candidate summary and the set of reference summaries. In Tab. \ref{tab:rouge}, comparison with three type of ROUGE-n is shown, ROUGE-$1$, ROUGE-$2$ and ROUGE-$3$. It can be seen that average recall is deceasing with increasing n-gram in ROUGE. The reason behind this behaviour is natural as ROUGE-$1$ compares on uni-gram basis in the candidate to reference corpus, which is word is word matching. ROUGE-$2$ compares on bi-gram basis, which is taking a set of two words at a time. However ROUGE-$3$ compares on tri-gram basis which is by considering $3$ words at a time. Since ROUGE-$1$, ROUGE-$2$, and ROUGE-$3$ use uni-gram, bi-gram and tri-gram comparisons respectively, the decreasing nature of average precision is natural. Machine generated descriptions has a fixed pattern for words to be used and the information to be displayed. However, human written descriptions can have any sequence and use of words and phrases. 

\subsubsection{BLEU}
BLEU metric analyses the co-occurrences of n-grams between a machine translation and human written sentence. The more the matches, the better is the candidate translation is. The score ranges from $0$ to $1$, where $0$ is the worst score and $1$ is the perfect match. In Tab. \ref{tab:BLEU}, we have given $4$ types of BLEU score, for $4$ values of n-gram. They first compute n-gram modified precision score $(p_n)$ by following equation
\begin{equation}
 p_n=\frac{\sum_{C\in \{Cand\}}{\sum_{gram-n\in C}{Count_{clip}(gram-n)}}}{\sum_{C'\in\{Cand\}}{\sum_{gram-n'\in C'}{Count(gram-n')}}}
\end{equation}
Where, $Count_{clip}$ limits the number of times a n-gram to be considered in a candidate ($Cand$) string.  
Then they computer the geometric mean of the modified precision $(p_n)$ using n-gram upto length $N$ and weights $W_n$ which sums up to $1$. A brevity penalty(BP) is used for longer candidate summaries and for spurious words in it, which is defined by the following equation:
%\begin{equation}
 \begin{equation}
    BP=
    \begin{cases}
      1, & \text{if}\ c>r \\
      e^{\frac{1-r}{c}}, & c\leq r
    \end{cases}
  \end{equation}
%\end{equation}
Where $c$ is the length of candidate summary and $r$ is the length of reference summary. Then BLEU score for corpus level given equal weights to all n-grams is evaluated by the following equation:
\begin{equation}
    BLEU=BP.exp^{\sum_{i=1}^{N}{W_n}log(p_n) }
\end{equation}
Here $W_n$ is the equally distributed weight in n-grams. E.g. in case of BLEU-$4$ the weights used are $\{(0.25),(0.25),(0.25),(0.25)\}$. The proposed dataset perform well on BLEU score as shown in Tab. \ref{tab:BLEU}. 
\subsubsection{METEOR}
METEOR is a metric used for evaluating machine generated summaries with human written summaries by checking the goodness of order of words in both. METEOR score is a combination of precision, recall and fragmentation (alignment) in the sentences. It is a harmonic mean of the uni-gram precision and uni-gram recall given an alignment, and calculated as:
\begin{equation}
    PN=\frac{1}{2}\left(\frac{no \;of\; chunks}{matched\; uni-grams}\right)
\end{equation}
%\begin{equation}
 %   F_{mean}=\frac{10PR}{R+9P}
%\end{equation}
\begin{equation}
    METEOR=\frac{10PR}{R+9P}(1-PN)
\end{equation}
Where $PN$ is the penalty imposed on the basis of larger number of chunks, $P$ is the uni-gram precision, $R$ is the uni-gram recall and METEOR is the final score obtained by multiplying the harmonic mean of uni-gram precision and uni-gram recall with penalty imposed. Table. \ref{tab:METEOR} shows the METEOR score we have obtained while experimenting on A-ROBIN dataset. It can be said that generated descriptions are very close to the human written descriptions. Also, it is clear that the descriptions collected in the A-ROBIN dataset are grammatically correct and close to the descriptions generated by the proximity based grammar model. 
\section{Qualitative results}
\label{sec:res}
In this section we describe the qualitative results. These result shows the generated descriptions for samples from A-ROBIN dataset, along with the navigation information in the narrative form.

\subsection{Examples of description Synthesis}
In SUGAMAN, rooms are labelled into one of the $5$ classes using the trained model (see Fig. \ref{fig:roomresult}). Room annotations and semantic information are stored in an XML file, which is parsed for description synthesis. Figure \ref{fig:resultdes} presents the resultant description for $3$ floor plan images. To facilitate the reader of the manuscript in understanding the description, we have made the following adjustments, (i) the first word of the first sentence about any room is in \textbf{bold} face, (ii) in the floor plan image, every room is highlighted with a different color and the same color is used to highlight the room name in the first sentence about the room, (iii) in the floor plan image,  the turning points are marked with 'T' and sequence of traversal of doors are marked with their respective numbers. There are two types of description synthesized for a given floor plan. The first kind of description is named as General description (GD), which contains information like name, area, global position in the floor plan, relative position of decors, and neighboring rooms in terms of its accessibility by a door is described for each room in the final output description, along with a room having a door opening to outside of the house is also described. The other one is Navigation description (NV), which contains navigation information from room to room avoiding obstacles. If a room has only one door, it is a dead end. Hence the navigating person will turn back. 

Figure. \ref{fig:resultdes}(a),(b),(c) are examples where the descriptions are successfully generated for the floor plan images. For  example, in Fig. \ref{fig:resultdes}(a), the GD correctly describes the number of rooms, there connectivity, as well as count and the arrangements of the decors inside each room. On the other hand, the NV part of the description guides the user to navigate to each room, starting from the entry. It can be observed that starting from entry door (labelled as $1$), the first obstacle to go to kitchen (as per the DFS navigation) is the sofa. Hence the user has to take a turn (marked as 'T') and then proceed. The directional information are obtained from the non-uniform binning technique discussed in Sec. \ref{sec:bin}. In Fig. \ref{fig:resultdes}(a), also the significance of ``backtracking'' can be understood. Once someone reaches the kitchen, then the next room to visit is the bedroom. Since there is no direct connection (as per Algo. 1, Line 5,  $AM_D(c_r,n_r)\not=1$). Thus the current room ($c_r$) is changed from kitchen to entry. However, the door from which the navigation should now be availed is $2$, even though the current room is the entry. The switching of the door index, as per door-to-door connectivity, is taken care by line 31-34 of Algo. 1. We have tested Algo. 1 for various configurations, and achieved correct results.

Figure. \ref{fig:resultdes}(d) shows a failure case. The reason behind this failure is that the decor recognition framework has miss-classified the large sink as sink and tub as large sofa. The NV part of this example is not shown here as the GD is incorrect. As a result the bathroom is labelled as kitchen. The limitations of the proposed framework are discussed next.
\begin{figure}[t]
    \centering
    \includegraphics[scale=0.29]{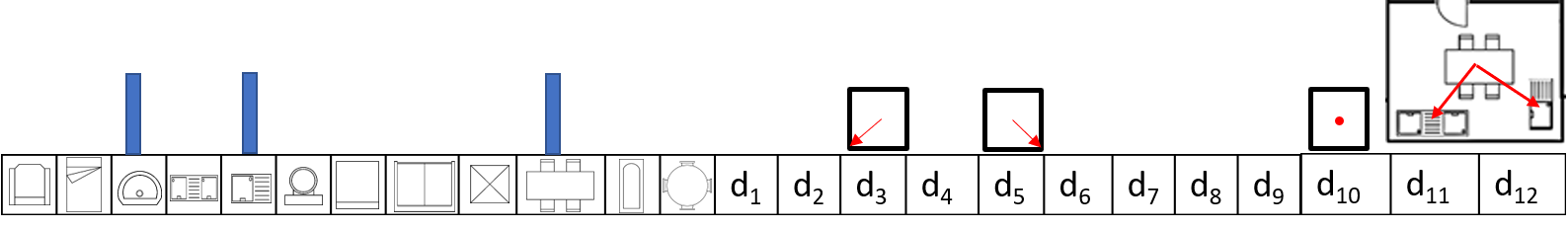}
    \caption{Incorrect creation of LOFD}
    \label{fig:limitation}
\end{figure}

\subsection{Limitations and future scope of LOFD feature}
The contents of LOFD feature vector highly depends on the decor classification. Inaccurate decor classification algorithm directly affects the accuracy of LOFD feature. Figure \ref{fig:limitation} depicts an incorrect creation of LOFD feature. In this, due to the morphological operation and joining of nearby blob, the twin sink is classified as large sink. As a result the classification model got confused for this room image and labelled the kitchen as bathroom. Hence, we require a highly accurate decor classification algorithm for a highly performing LOFD. Reducing its dependency over decor classification or a better decor classification algorithm will further improve its accuracy. Moreover, the descriptions collected through Google form at present is scripted.However, in reality there can be ambiguity in terms of how a floor plan is described, there may be incomplete sentences, information about a given room may not be at the same place, etc. A learning model that can collate all these information and generates a single description to be used for experiment purpose forms a unique scope for future work.

\section{Conclusion and Future work}
\label{sec:conclusion}
The primary objective of this work is to generate description and navigation information from floor plan images for the visually impaired. We have proposed a novel LOFD feature for automatic room label learning. A proximity based grammar model is also proposed used to synthesize the description. Navigation information in the form of narration is also generated by the proposed algorithm. We have also propsed a novel description dataset, A-ROBIN and made it publicly available for the DAR community. In future we are planning to introduce machine learning methods, NLP techniques and improved feature to generated description using the human generated descriptions present in dataset to further improve the performance and generalize the proposed system.

\end{document}